\documentclass{ieee-ojvt}
\usepackage{cite}
\usepackage{textcomp}
\usepackage{amsmath,amssymb,amsfonts}
\usepackage{algorithmic}
\usepackage{stfloats}
\usepackage{array}
\usepackage{verbatim}
\usepackage{graphicx,color}
\usepackage{textcomp}
\usepackage{booktabs}
\usepackage{multirow}
\usepackage{balance}
\usepackage{pifont}
\usepackage{graphicx}
\usepackage{tabularray}
\usepackage{tabularx}
\usepackage{ragged2e}
\usepackage[switch]{lineno}
\usepackage{xcolor}
\usepackage{graphicx} 
\usepackage{multirow}
\usepackage{xcolor}
\usepackage{array}
\usepackage{rotating}
\usepackage{makecell}
\usepackage{float}
\usepackage{amsmath}
\usepackage{hyperref}
\usepackage{algorithm}
\usepackage{algorithmic}
\usepackage{amsmath, amsfonts}

\usepackage{float}
\usepackage{placeins}
\usepackage[T1]{fontenc}
\def\BibTeX{{\rm B\kern-.05em{\sc i\kern-.025em b}\kern-.08em
    T\kern-.1667em\lower.7ex\hbox{E}\kern-.125emX}}
\AtBeginDocument{\definecolor{ojcolor}{cmyk}{0.93,0.59,0.15,0.02}}
\def\OJlogo{\vspace{-12pt}\includegraphics[height=24pt]{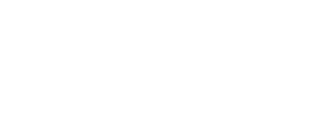}}

\begin{document}
\receiveddate{XX March, 2026}

\title{Learnable Quantum Efficiency Filters for Urban Hyperspectral Segmentation}

\author{
    Imad Ali Shah\textsuperscript{1,2},
    Jiarong Li\textsuperscript{1,2},
    Ethan Delaney\textsuperscript{1,2},
    Enda Ward\textsuperscript{3},
    Martin Glavin\textsuperscript{1,2},
    Edward Jones\textsuperscript{1,2} and 
    Brian Deegan\textsuperscript{1,2}
}

\affil{School of Engineering, University of Galway, Ireland}
\affil{Ryan Institute, University of Galway, Ireland}
\affil{Valeo Vision Systems, Tuam, Ireland}

\corresp{CORRESPONDING AUTHOR: Imad Ali Shah (e-mail: i.shah2@universityofgalway.ie).}
\authornote{
This work was supported, in part, by Taighde Éireann -- Research Ireland grants 13/RC/2094\_P2 and 18/SP/5942, and co-funded under the European Regional Development Fund through the Southern and Eastern Regional Operational Programme to Lero -- the Research Ireland Centre for Software (www.lero.ie), and by Valeo Vision Systems.
}
\markboth{Learnable Quantum Efficiency Filters for Urban Hyperspectral Segmentation}{Shah \textit{et al.}}
\begin{abstract}
Hyperspectral sensing provides rich spectral information for scene understanding in urban driving, but its high dimensionality poses challenges for interpretation and efficient learning. We introduce Learnable Quantum Efficiency (LQE), a physics-inspired, interpretable dimensionality reduction (DR) method that parameterizes smooth high-order spectral response functions that emulate plausible sensor quantum efficiency curves. Unlike conventional methods or unconstrained learnable layers, LQE enforces physically motivated constraints, including a single dominant peak, smooth responses, and bounded bandwidth. This formulation yields a compact spectral representation that preserves discriminative information while remaining fully differentiable and end-to-end trainable within semantic segmentation models (SSMs). We conduct systematic evaluations across three publicly available multi-class hyperspectral urban driving datasets, comparing LQE against six conventional and seven learnable baseline DR methods across six SSMs. Averaged across all SSMs and configurations, LQE achieves the highest average mIoU, improving over conventional methods by 2.45\%, 0.45\%, and 1.04\%, and over learnable methods by 1.18\%, 1.56\%, and 0.81\% on HyKo, HSI-Drive, and Hyperspectral City, respectively. LQE maintains strong parameter efficiency (12--36 parameters compared to 51--22K for competing learnable approaches) and competitive inference latency. Ablation studies show that low-order configurations are optimal, while the learned spectral filters converge to dataset-intrinsic wavelength patterns. These results demonstrate that physics-informed spectral learning can improve both performance and interpretability, providing a principled bridge between hyperspectral perception and data-driven multispectral sensor design for automotive vision systems.
\end{abstract}

\maketitle
\section{Introduction}
\label{sec:intro}

\begin{figure}[H]
   \includegraphics[width=0.98\columnwidth]{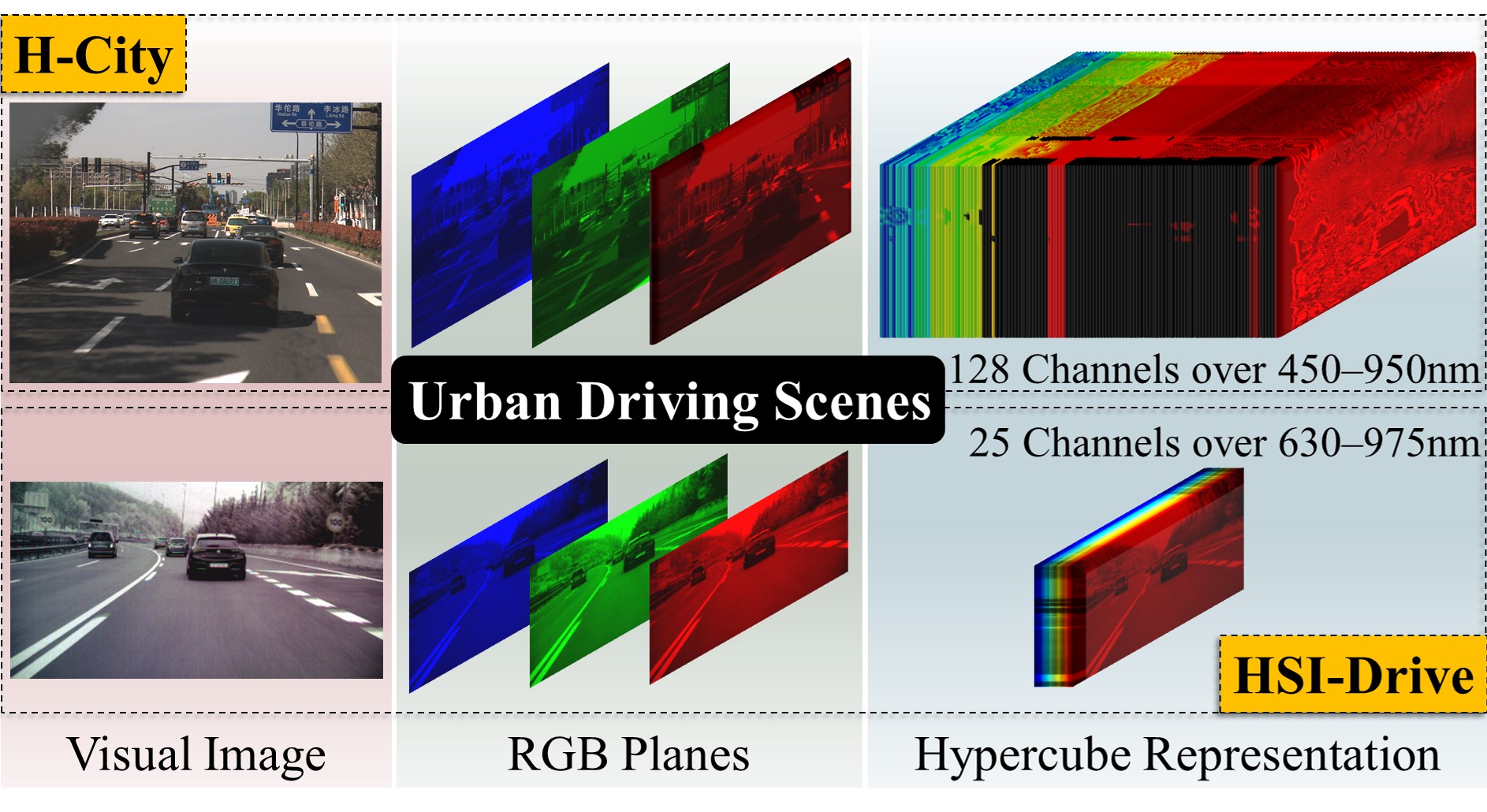}
   \vspace{-0.5em}
   \caption{Comparison of visual images, RGB Planes, and hypercube.}
   \label{fig:RGBvsHSI_Comparison}
\end{figure}

\IEEEPARstart{S}{afe} autonomous driving requires robust perception, yet widely deployed RGB cameras are susceptible to metamerism~\cite{thornton1998strong}. Hyperspectral imaging~(HSI) mitigates this limitation by capturing dense spectral signatures across hundreds of wavelengths~\cite {lu2020recent}, enabling improved material discrimination for road surface assessment~\cite{ozdemir2020neural,abdellatif2020pavement} and pedestrian detection~\cite{li2025hyperspectralvsrgbpedestrian}. However, this spectral richness introduces high dimensionality, as illustrated in Fig.~\ref{fig:RGBvsHSI_Comparison} with samples from Hyperspectral City~(H-City)~\cite{shen4560035urban} and HSI-Drive~\cite{gutierrez2023hsi} datasets, creating a challenge for efficient learning.

\begin{figure*}[t!]
  \centering
  \includegraphics[width=\textwidth]{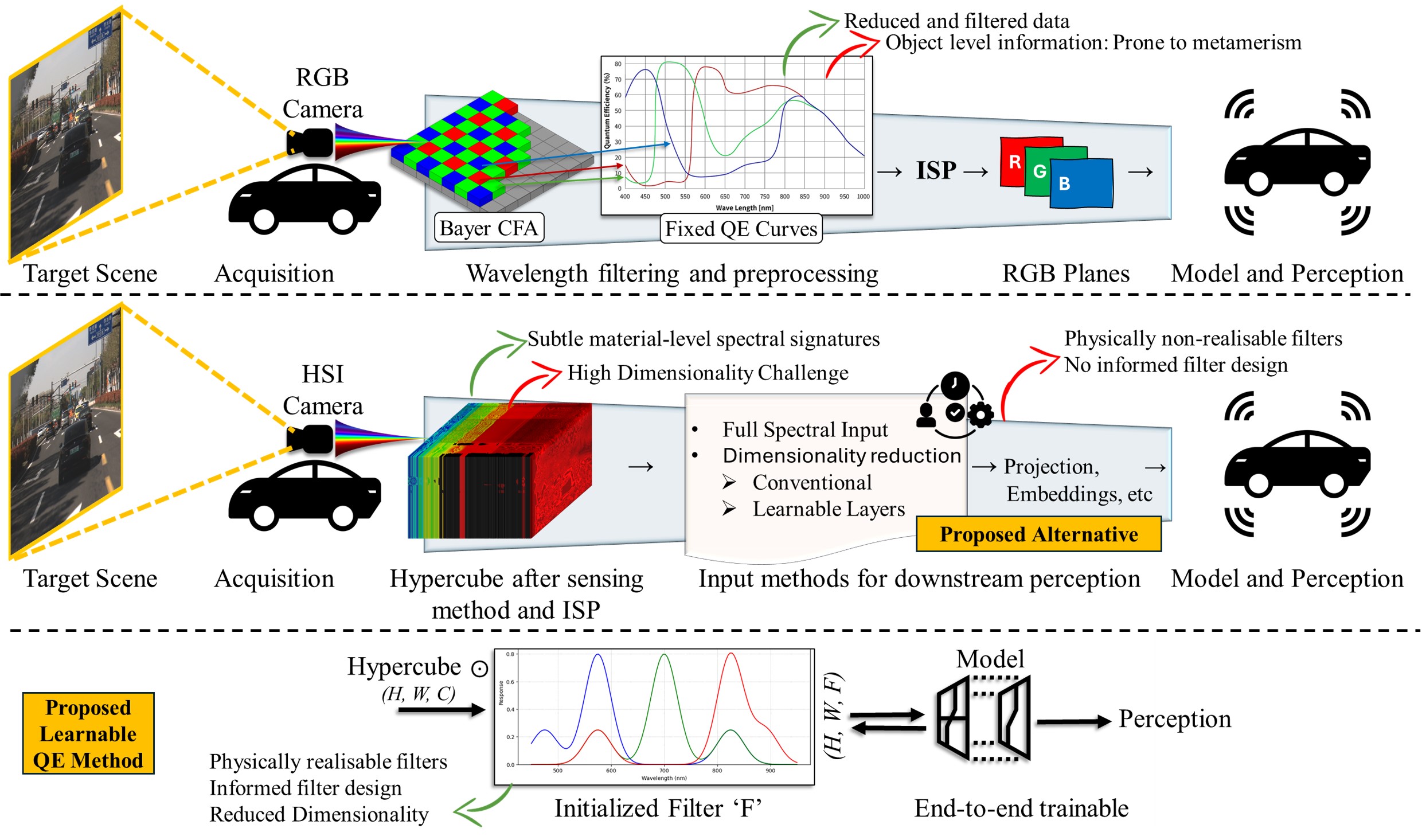}
  \vspace{-1.5em}
  \caption{Overview of conventional imaging pipelines and the proposed learnable quantum efficiency~(LQE) framework. (Top) Standard RGB systems employ fixed Bayer CFA, producing low-dimensional data that are computationally efficient but susceptible to metamerism and loss of material-specific information. (Middle) HSI captures dense spectra but incurs substantial computational and memory overhead due to high dimensionality, requiring DR. (Bottom) The proposed LQE approach introduces continuously differentiable, high-order QE filters that are jointly optimized with the downstream perception model and optically plausible. This framework learns a compact, task-adaptive spectral projection of the hypercube, mitigating the computational burden of HSI while preserving discriminative spectral content. }
  \label{fig:Fig_paperIntroduction}
\end{figure*}

As shown in Fig.~\ref{fig:Fig_paperIntroduction}, conventional RGB cameras inherently perform spectral dimensionality reduction~(DR) through fixed color filter arrays~(CFAs). These CFA come with predetermined quantum efficiency~(QE) curves, which are wavelength-dependent photon-to-electron conversion probabilities, designed to capture specific but broad spectral regions. While effective and remaining highly successful for general-purpose imaging tasks, including object recognition and lane keeping~\cite{campbell2018sensor}, these filters are pre-designed independently of downstream perception objectives and inherit limitations such as illumination sensitivity and metamerism~\cite{thornton1998strong}. HSI, on the other hand, can overcome this RGB limitation and effectively reduces metameric-ambiguity by leveraging dense spectra to detect vulnerable road users~\cite{li2025hyperspectralvsrgbpedestrian} in visually complex situations.

A typical HSI cube, commonly known as a hypercube ($x, y, \lambda$: wavelength), can have 25--128 spectral bands~\cite{gutierrez2023hsi,shen4560035urban} compared to 3-channel RGB, requires over 8--42$\times$ more data throughput~\cite{shah2025hyperspectral}. Moreover, its spectral redundancy~\cite{arneal2015spectral} imposes significant computational burdens on optimizing downstream vision tasks, and fine spectral sampling reduces per-band photon capture, degrading signal-to-noise ratios in low-light conditions~\cite{zhang2023snapshot}. In addition, the current availability, form factors, and cost of HSI cameras further limit automotive integration~\cite{shah2025hyperspectral}. These constraints motivate task-aware spectral DR that preserves discriminative information while enabling efficient deployment and interpretability for further research to inform multispectral sensor design.

\begin{figure}[t]
  \centering
   \includegraphics[width=\columnwidth]{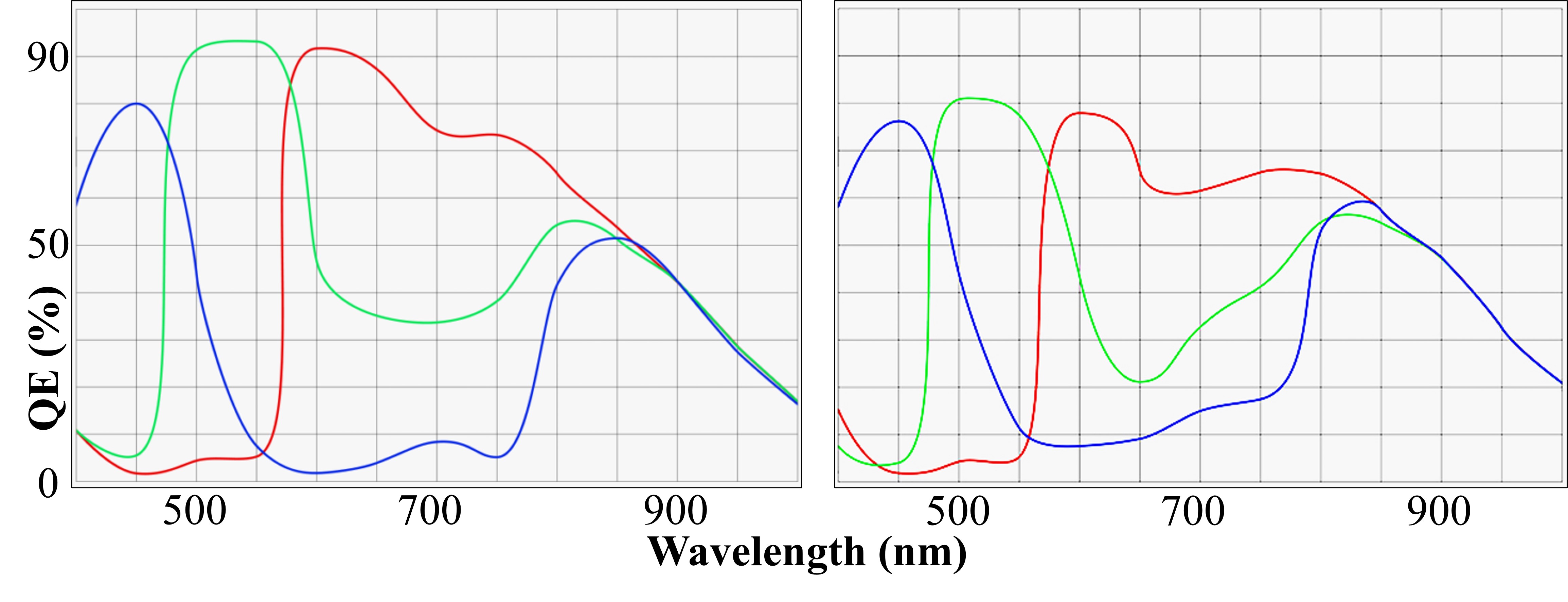}
   \vspace{-1.75em}
   \caption{Example QE curves of the ZWO ASI662MC (left) and ASI462MC (right) sensors' CFA~\cite{ZWOAstro_ASI662MC_ASI462MC}, exhibiting a dominant spectral peak, smooth spectral response, and bounded bandwidth.}
   \label{fig:lqe_Inspiration_Figure}
\end{figure}

Existing HSI DR methods are either classical, offering interpretability but no task adaptation, or recent learnable approaches, enabling end-to-end optimization but typically operating as unconstrained black-box projections. Neither of these approaches enforces physical constraints consistent with real optical filter responses. 
Instead of viewing DR as a feature embedding problem, we reinterpret it as a constrained optical filter design problem. Specifically, we parameterize each spectral response as a smooth, bounded function, preventing arbitrary or physically implausible transformations that are inspired by the QE curves of physical CFAs, examples of which are illustrated in Fig.~\ref{fig:lqe_Inspiration_Figure}. We introduce differentiable high-order~(multi-peak) QE filters that emulate plausible sensor QE curves while remaining compatible with gradient-based optimization within modern deep learning~(DL) frameworks.

Compared to classical and unconstrained learnable DR methods, LQE enables end-to-end optimization while producing interpretable spectral response curves. The learned filters highlight discriminative wavelength regions for urban scene understanding, potentially informing future multispectral sensor design research. Our primary contributions include:
\begin{enumerate}
    \item We introduce a differentiable high-order QE filter parameterization using asymmetric Gaussian basis functions with physics-inspired constraints~(peak dominance, centroid separation, and bandwidth bounds), enabling end-to-end optimized spectral reduction.
    \item We conduct extensive evaluation on all publicly available HSI urban driving datasets at the time of this work, across six classical and seven learnable DR methods, using six segmentation architectures.
    \item Through systematic ablation (sub-peaks per filter), we show that low-order configurations are optimal, with learned filters exhibiting architecture-agnostic convergence to dataset-intrinsic spectral patterns across tested architectures.
\end{enumerate}

The rest of the paper includes Section~\ref{sec:relatedWork}, discussing related work; Section~\ref{sec:methodology}, detailing the methodology; Section~\ref{sec:ExperimentationAndResults}, providing experimental results; and Section~\ref{sec:Conclusion}, concluding the work.

\section{Related Work}\label{sec:relatedWork}
\subsection{HSI for Scene Understanding}
HSI has recently gained attention in ADAS and autonomous driving perception for its ability to capture fine-grained spectral signatures that improve robustness under visually 
challenging conditions~\cite{shah2025hyperspectral,li2025hyperspectralvsrgbpedestrian}. Unlike RGB sensors, HSI preserves discriminative spectral detail that benefits safety-critical tasks, including road and pavement assessment~\cite{valme2024road,abdellatif2020pavement}, pedestrian detection~\cite{li2025hyperspectralvsrgbpedestrian}, and material classification~\cite{li2025csnrjmimbasedspectral}, effectively mitigating metameric-ambiguity inherent to broad-spectrum RGB imaging. Publicly available multi-class HSI datasets for urban driving, including HyKo~\cite{winkens2017hyko}, HSI-Drive~\cite{gutierrez2023hsi}, and H-City~\cite{shen4560035urban}, summarized in Table~\ref{tab:hsi_datasets}, have enabled early exploration of DL for HSI modality.
Critically, no prior HSI urban perception work has coupled spectral DR with continuously differentiable, physically plausible QE filter constraints across VIS-NIR ranges for semantic segmentation.

\begin{table}[h]
\caption{All available multiclass HSI datasets for urban scenes}
\label{tab:hsi_datasets}
\resizebox{0.489\textwidth}{!}{%
\begin{tabular}{l c r r c r}
\toprule
\textbf{Datasets} & \textbf{Spatial} & \textbf{Spectral} & \textbf{Classes} & \textbf{Range} & \textbf{No of} \\
& \textbf{Dimension} & \textbf{Channels} & & \textbf{(nm)} & \textbf{Images} \\
\midrule
HyKo-VIS~\cite{winkens2017hyko} & 254x512 & 15 & 10 & 470-630 & 163 \\
HSI-Drive~\cite{gutierrez2023hsi} & 209x416 & 25 & 9 & 600-975 & 752 \\
H-City\textsuperscript{*}~\cite{shen4560035urban} & 1422x1889 & 128 & 19 & 450-950 & 1,330 \\
\bottomrule
\end{tabular}
}

\footnotesize \vspace{1.75pt} \textsuperscript{*} The only multi-class HSI dataset with co-registered RGB
\end{table}

\subsection{Dimensionality Reduction~(DR)}
Existing DR methods can be broadly divided into classical and learnable approaches; however, their suitability for physically grounded HSI urban driving scenarios remains underexplored:

\textbf{Classical Methods.} Traditional analytical techniques, including Principal Component Analysis~(PCA)~\cite{Pearson01111901} and Independent Component Analysis~(ICA)~\cite{hyvarinen2000independent}, optimize statistical criteria such as data variance along orthogonal projections and statistical independence. While Linear Discriminant Analysis~(LDA)~\cite{loog_dimensionality_2005} maximizes inter-class separability in a supervised setting. Manifold learning methods, including Locally Linear Embedding~(LLE)~\cite{roweis2000nonlinear} and Isomap~\cite{tenenbaum2000global}, preserve local neighborhood structures in reduced spaces, while Maximum Noise Fraction~(MNF)~\cite{green1988transformation} and Non-Negative Matrix Factorization~(NMF)~\cite{lee1999learning} aim at noise suppression and part-based decompositions, respectively. Although these methods offer computational efficiency and statistical interpretability, they are distinguished from \textit{optical} interpretability: PCA or ICA projections lack physical correspondence to plausible sensor QE curves, offering no actionable spectral guidance. Furthermore, classical approaches are by design task-agnostic, optimizing statistical objectives decoupled from the downstream perception objectives.

\textbf{Learnable End-to-End DR.} Modern DL-based approaches integrate DR directly into perception networks, making them end-to-end optimizable. Convolutional~(Conv) bottlenecks, such as a 1$\times$1 spatial Conv~(1$\times$1Conv)~\cite{he2016deep,li2016hyperspectral} and autoencoders~(AE)~\cite{chen2014deep}, project high-dimensional spectral data to low-dimensional latent spaces. Channel attention modules, including Conv Block Attention Module~(CBAM)~\cite{woo2018cbam}, Coordinate Attention~(CA)~\cite{dang2021coordinate}, and Squeeze-and-Excitation~(SE)~\cite{hu2018squeeze} dynamically weight spectral bands to retain informative wavelengths. While these approaches achieve strong task performance, their learned transformations are unconstrained black-box projections that lack physical interpretability.

\subsection{Hardware-Aware Learning}
A hardware-aware research direction seeks to unify data-driven optimization with physical imaging models. Previous works on spatial-placement CFA design patterns~\cite{chakrabarti2016learning}, differentiable image signal processing pipelines~\cite{tseng2021differentiable}, wavelength-aware Conv2D~\cite{varga2023wavelength}, and hardware-software co-design~\cite{zamir2020cycleisp} demonstrated that physical imaging stages can be jointly optimized with perception objectives.

However, these approaches share important limitations: either (1)~they optimize spatial multiplexing patterns rather than continuous spectral response functions across extended visible-to-near-infrared (VIS-NIR) ranges; or (2)~they do not constrain learned filters to the smooth, bounded, multi-peak structure consistent with real sensor spectral response behavior. None of these approaches enforces smooth, bounded, multi-peak QE constraints across extended spectral ranges, which is the specific gap our proposed LQE addresses. These requirements are necessary for translating existing hyperspectral constraints into informed spectral information for future efficient multispectral approaches.

\subsection{Position of Contribution}
No existing work or method jointly satisfies task-optimal end-to-end DR, physical plausibility of learned spectral response filters, and optical interpretability for further data-driven sensor design research. This is the first framework, particularly within the HSI urban driving domain, to jointly satisfy all these requirements. LQE addresses this gap by parameterizing each spectral response as a continuously differentiable, high-order (multi-peak) function across VIS-NIR wavelengths, reframing DR as a constrained optical filter design problem. 

\section{Methodology}
\label{sec:methodology}
\subsection{Architecture Overview}
We propose a differentiable multi-peak QE filter bank for learning compact, physically interpretable spectral representations from HSI data. While QE denotes the photon-to-electron conversion efficiency, we use the term in this work to refer to the wavelength-dependent spectral response function of a CFA element, which QE curves characterize in practice.

The LQE architecture transforms an input hypercube $\mathbf{X} \in \mathbb{R}^{B \times C \times H \times W}$ into a reduced representation $\mathbf{Y} \in \mathbb{R}^{B \times F \times H \times W}$, 
where $B$ is the batch size, $C$ the number of input spectral channels, 
$H \times W$ the spatial dimensions, and $F \ll C$ the reduced number of 
learned filters. Each learned filter emulates the spectral response characteristics of a physical CFA element.

\subsection{Multi-Peak Filter Formulation}

Each filter $Q_f$ in the bank is parameterized as a weighted sum of $P$ asymmetric Gaussian functions, enabling the representation of complex, multimodal spectral responses. The model operates on a normalized wavelength coordinate $\tilde{\lambda} \in [0,1]$, defined as:
\begin{equation}
\tilde{\lambda} = \frac{\lambda - \lambda_{\text{start}}}{\lambda_{\text{end}} - \lambda_{\text{start}}}
\end{equation}

\noindent where $\lambda \in [\lambda_{\text{start}}, \lambda_{\text{end}}]$ represents the wavelength range specific to the data set.

For $p$ peaks in each filter, we maintain four learnable parameters: Centroid $c_{p,f} \in [0,1]$ defines the peak position; Log-bandwidth $\ell_{p,f} \in \mathbb{R}$ controls width through $\beta_{p,f} = \exp(\ell_{p,f})$; Amplitude $\alpha_{p,f} \in \mathbb{R}$ mapped to $a_{p,f} = \sigma(\alpha_{p,f}) \in (0,1)$; and skewness $\gamma_{p,f} \in \mathbb{R}$ introduces asymmetry. 
The standardized distance from the peak centroid is:
\begin{equation}
x_{p,f}(\tilde{\lambda}) = \frac{\tilde{\lambda} - c_{p,f}}{\beta_{p,f}}
\label{Eq.2}
\end{equation}

To model asymmetries observed in real CFA spectral responses (Fig.~\ref{fig:lqe_Inspiration_Figure}), we introduce a skewness-controlled transformation of the standardized distance through $s_{p,f} = 0.5\tanh(\gamma_{p,f})$, designed to preserve the centroid parameterization $c_{p,f}$. QE curves are empirically asymmetric due to material and optical properties, making a symmetric Gaussian basis insufficient. Our formulation applies a bounded, smooth $\tanh$-based modulation centered at $x_{p,f}=0$, ensuring that the response is anchored at the centroid while modifying its shape, in particular its skewness. This reduces coupling between asymmetry and centroid location, maintaining interpretability by making centroid and skewness largely decoupled. Alternative distributions, such as log-normal or skew-normal, typically introduce stronger coupling between skewness and the effective mean, while other approaches, including Voigt profiles or piecewise asymmetric functions, either increase parameterization or constrain continuous differentiability. The proposed formulation, therefore, achieves a balance between QE fidelity, differentiability, and parametric efficiency.
\begin{equation}
x_{p,f}^{\text{skew}}(\tilde{\lambda}) = x_{p,f}(\tilde{\lambda})\left(1 + s_{p,f} \cdot \text{tanh}(x_{p,f}(\tilde{\lambda}))\right)
\label{Eq.3}
\end{equation}

\noindent The $p$-th peak response is:
\begin{equation}
g_{p,f}(\tilde{\lambda}) = a_{p,f} \exp\left(-\frac{1}{2}\left(x_{p,f}^{\text{skew}}(\tilde{\lambda})\right)^2\right)
\label{Eq.4}
\end{equation}

\noindent The normalized filter response over all wavelengths:
\begin{equation}
Q_f(\tilde{\lambda}) = \frac{\sum_{p=1}^{P} g_{p,f}(\tilde{\lambda})}{\max_{\tilde{\lambda}}\left(\sum_{p=1}^{P} g_{p,f}(\tilde{\lambda})\right) + \varepsilon}
\label{Eq.5}
\end{equation}
where $\varepsilon = 10^{-8}$ prevents numerical instability.

\subsection{Filter Bank Operation}

Given hypercube input $\mathbf{X}$ with channel wavelengths $\{\tilde{\lambda}_c\}_{c=1}^{C}$, each filter $f$ performs a differentiable spectral integration:
\begin{equation}
\mathbf{Y}_f(b,h,w) = \sum_{c=1}^{C} Q_f(\tilde{\lambda}_c) \cdot \mathbf{X}(b,c,h,w)
\label{Eq.6}
\end{equation}

The complete reduced tensor is obtained as $\mathbf{Y} = [\mathbf{Y}_1, \dots, \mathbf{Y}_F]$. 

\subsection{End-to-End Optimization}
The four learnable parameters ($c_{p,f}$, $\ell_{p,f}$, $\alpha_{p,f}$ and $\gamma_{p,f}$) per peak $p$ are optimized end-to-end through backpropagation. Gradients flow from the downstream SSM through the differentiable spectral integration (Eq.~\ref{Eq.6}) back to all filter parameters: $\alpha_{p,f}$ (Eq.~\ref{Eq.4}) $\rightarrow$ $\gamma_{p,f}$ (Eq.~\ref{Eq.3}) $\rightarrow$ $c_{p,f}$ and $\ell_{p,f}$ in Eq.~\ref{Eq.2}, respectively. This is possible because the filter responses $Q_f$ (Eq.~\ref{Eq.5}) are continuously differentiable with respect to all four parameters at every wavelength coordinate.

\subsection{Physics-Inspired Regularization}

To ensure physically plausible and discriminative spectral response filters, we employ three regularization terms:

\textbf{Peak Dominance Loss}: Encourages a single dominant spectral lobe per filter:

\begin{equation}
\mathcal{L}_{\text{dom}}^{(f)} = \text{ReLU}\!\left(\frac{a_{\text{2nd}}^{(f)}}{a_{p_f^*}^{(f)} + \varepsilon} - r_{\text{max}}\right)    
\end{equation}

\begin{equation}
\mathcal{L}_{\text{dom}} = \tfrac{1}{F}\!\sum_{f=1}^{F}\! \mathcal{L}_{\text{dom}}^{(f)}
\label{Eq.8}
\end{equation}

where $a_{p_f^*}^{(f)}$ is the maximum peak amplitude, $a_{\text{2nd}}^{(f)}$ is the largest secondary peak amplitude, and $r_{\text{max}}$ (typically 0.3) controls their relative contribution.

\textbf{Centroid Separation Loss}: Promotes spectral diversity by enforcing minimum spacing $d_{\text{min}}$ between the primary peak centroids $c_f^*$:
\begin{equation}
    \mathcal{L}_{\text{sep}} = \frac{1}{F^2}\sum_{f=1}^{F}\sum_{k \neq f} \text{ReLU}(d_{\text{min}} - |c_f^* - c_k^*|)
    \label{Eq.9}
\end{equation}

\textbf{Bandwidth Regularization}: Constrains each filter’s bandwidth $\beta_f^*$ to physically plausible limits $(\beta_{\text{min}}: 0.03, \beta_{\text{max}}: 0.25]$, where $\beta_{\text{min}}$ produces 4.8--15nm bandwidth for HSI datasets in Table~\ref{tab:hsi_datasets}, supporting the spectral resolution of 4--15nm of currently available snapshot HSI Cameras~\cite{shah2025hyperspectral}:
\begin{equation}
\mathcal{L}_{\text{bw}} = \frac{1}{F}\sum_{f=1}^{F} \left[\text{ReLU}(\beta_{\text{min}} - \beta_f^*) + \text{ReLU}(\beta_f^* - \beta_{\text{max}})\right]
\label{Eq.10}
\end{equation}

The total regularization objective $\mathcal{L}_{\text{reg}}$ is:
\begin{equation}
    \mathcal{L}_{\text{reg}} = \mathcal{L}_{\text{dom}} + \mathcal{L}_{\text{sep}} + \mathcal{L}_{\text{bw}}.
    \label{Eq.11}
\end{equation}

The total objective loss $\mathcal{L}_{\text{total}}$:
\begin{equation}
    \mathcal{L}_{\text{total}} = \mathcal{L}_{\text{seg}}(\mathbf{Y}_{\text{pred}}, \mathbf{Y}_{\text{target}}) + \lambda_{\text{reg}}\mathcal{L}_{\text{reg}}; \quad \lambda_{\text{reg}}=0.1
    \label{Eq.12}
\end{equation}

where segmentation loss $\mathcal{L}_{\text{seg}}$ is the primary driver of task performance in SSMs, and the regularization term $\mathcal{L}_{\text{reg}}$ serves as physics-inspired constraints.

\subsection{Proposed LQE}
\label{subsec:ProposedLQE}

Algorithm~\ref{alg:lqe} presents the complete computational procedure for the LQE filter bank, including initialization, forward pass, and regularization loss computation.

\begin{algorithm}[h!]
\caption{LQE Filter Bank}
\begin{algorithmic}[1]
\REQUIRE Hypercube $\mathbf{X} \in \mathbb{R}^{B \times C \times H \times W}$, wavelength range $[\lambda_{\text{start}}, \lambda_{\text{end}}]$, number of filters $F$, peaks per filter $P$
\ENSURE Reduced representation $\mathbf{Y} \in \mathbb{R}^{B \times F \times H \times W}$

\STATE \textbf{Initialize:} For each filter $f \in \{1, \dots, F\}$:
\STATE \quad Centroids: $\mathbf{c}_f \sim \text{Uniform}(0.1, 0.9) + \mathcal{N}(0, 0.05^2)$ \COMMENT{Uniform + Gaussian noise}
\STATE \quad Log-bandwidth: $\ell_{p,f} \gets \log(0.05 + 0.02 \cdot \text{rand}())$
\STATE \quad Amplitudes: $\alpha_{p,f} \sim \mathcal{N}(0, 0.5^2)$
\STATE \quad Skewness: $\gamma_{p,f} \gets 0$

\STATE \textbf{Normalize wavelengths:} \\ $\tilde{\lambda}_c = \frac{\lambda_c - \lambda_{\text{start}}}{\lambda_{\text{end}} - \lambda_{\text{start}}}$, $\forall c \in \{1, \dots, C\}$

\FOR{each filter $f = 1$ to $F$}
    \FOR{each peak $p = 1$ to $P$}
        \STATE Compute bandwidth: $\beta_{p,f} = \exp(\ell_{p,f})$
        \STATE Compute amplitude: $a_{p,f} = \sigma(\alpha_{p,f})$
        \STATE Compute skewness: $s_{p,f} = 0.5 \cdot \tanh(\gamma_{p,f})$
        \STATE Standardized distance: $x_{p,f}(\tilde{\lambda}_c) = \frac{\tilde{\lambda}_c - c_{p,f}}{\beta_{p,f}}$
        \STATE Skewed distance: \\ $x_{p,f}^{\text{skew}}(\tilde{\lambda}_c) = x_{p,f}(\tilde{\lambda}_c) \cdot [1 + s_{p,f} \cdot \tanh(x_{p,f}(\tilde{\lambda}_c))]$
        \STATE Peak response: \\ $g_{p,f}(\tilde{\lambda}_c) = a_{p,f} \cdot \exp\left(-\frac{1}{2}[x_{p,f}^{\text{skew}}(\tilde{\lambda}_c)]^2\right)$
    \ENDFOR
    \STATE Aggregate peaks: $Q_f(\tilde{\lambda}_c) = \frac{\sum_{p=1}^{P} g_{p,f}(\tilde{\lambda}_c)}{\max_c \sum_{p=1}^{P} g_{p,f}(\tilde{\lambda}_c) + \epsilon}$
    \STATE Apply filter: $\mathbf{Y}_f = \sum_{c=1}^{C} Q_f(\tilde{\lambda}_c) \cdot \mathbf{X}_{:,c,:,:}$ \quad \COMMENT{Weighted spectral integration}
\ENDFOR

\STATE \textbf{Regularization losses:}
\STATE Peak dominance: \\ $\mathcal{L}_{\text{dom}} = \frac{1}{F}\sum_{f=1}^{F} \text{ReLU}\left(\frac{a_{2\text{nd}}^{(f)}}{a_{p^*_f}^{(f)} + \epsilon} - r_{\text{max}}\right)$
\STATE Centroid separation: \\ $\mathcal{L}_{\text{sep}} = \frac{1}{F^2}\sum_{f=1}^{F}\sum_{k \neq f} \text{ReLU}(d_{\text{min}} - |c_{f}^* - c_{k}^*|)$
\STATE Bandwidth bounds: \\ $\mathcal{L}_{\text{bw}} = \frac{1}{F}\sum_{f=1}^{F} [\text{ReLU}(\beta_{\text{min}} - \beta_f^*) + \text{ReLU}(\beta_f^* - \beta_{\text{max}})]$

\RETURN $\mathbf{Y} = [\mathbf{Y}_1, \dots, \mathbf{Y}_F]$, $\mathcal{L}_{\text{reg}} = \mathcal{L}_{\text{dom}} + \mathcal{L}_{\text{sep}} + \mathcal{L}_{\text{bw}}$
\end{algorithmic}
\label{alg:lqe}
\end{algorithm}

\begin{figure}[!t]
    \centering
    \includegraphics[width=0.45\textwidth]{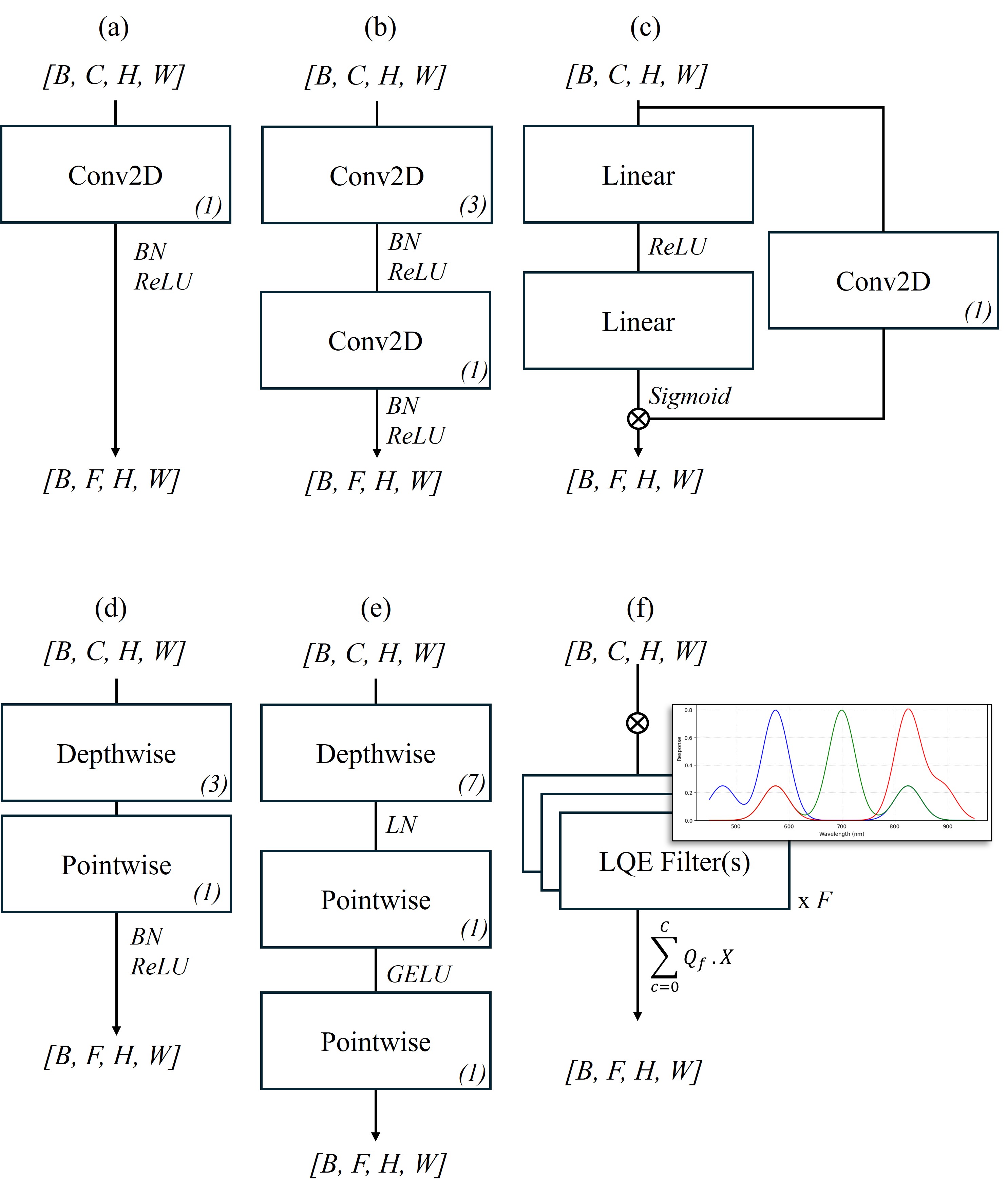}
    \caption{Comparison of learnable DR layers. All methods transform input hypercubes from $C$ spectral channels to $F$ reduced channels~($F \ll C$) while preserving spatial dimensions $H \times W$. (a)~1×1Conv: Standard Conv with batch normalization and ReLU. (b)~AE. (c)~seAttn: Gating mechanism with SE channel reweighting. (d)~DSC: Depthwise separable convolution. (e)~CN: ConvNeXt-style block. Others include original CBAM and ECA style layers. (f)~Proposed LQE method parameterizes spectral response filters as high-order Gaussian basis functions $Q_f$. Numbers in parentheses indicate:  kernel sizes. BN: Batch Normalization, and LN: Layer Normalization.} 
    \label{fig:Suppl_Architectures}
\end{figure}

\subsection{Implementation Details}
The proposed LQE contains $4PF$ learnable parameters in total, for $F$-filters with $P$-peaks each. \textit{Initialization}: Filter centroids are initialized uniformly across the spectral range with small random perturbations to encourage diversity. Bandwidths are initialized to moderate values ($\beta_p \!\approx\! 0.05$--$0.07$), amplitudes randomly in $(0,1)$, and skewness parameters to zero, yielding initially symmetric responses. \textit{Architecture Integration}: The QE filter bank replaces conventional DR preprocessing layers, operating directly on hyperspectral inputs to produce $F$ reduced channels. The resulting representation is fed into downstream semantic segmentation models~(SSM). For consistency, we evaluate all SSMs using the Segmentation Models Pytorch~(SMP)~\footnote{Available and accessed at \url{https://github.com/qubvel-org/segmentation\_models.pytorch}} library.

\section{Experimentation and Results}
\label{sec:ExperimentationAndResults}

\subsection{Experimental Setup}
\subsubsection{Datasets and Segmentation Models}
Three public multi-class HSI datasets for urban scenes: HyKo~\cite{winkens2017hyko}, HSI-Drive~\cite{gutierrez2023hsi}, and H-City~\cite{shen4560035urban} (Details in Table~\ref{tab:hsi_datasets}), were used for evaluation. To ensure architecture-agnostic evaluation, each DR method was tested across six SSMs: U-Net~(EfficientNet-B0), U-Net++~(EfficientNet-B3), DeepLabV3+~(ResNet50), PSPNet~(ResNet50), FPN~(ResNeXt50-32x4d), and SegFormer~(MiT-B3), implemented via the SMP library. All reported metrics are averaged across these six architectures.

\subsubsection{Baseline DR Methods: Classical and Learnable}
The proposed LQE approach was compared against six conventional DR techniques~(PCA~\cite{Pearson01111901}, ICA~\cite{hyvarinen2000independent}, NMF~\cite{lee1999learning}, MNF~\cite{green1988transformation}, LLE~\cite{roweis2000nonlinear}, and Isomap~\cite{tenenbaum2000global}), as well as seven lightweight learnable layers: 1x1Conv~\cite{he2016deep,li2016hyperspectral}, AE~\cite{ranzato2007unsupervised,masci2011stacked}, Efficient Channel Attention~(ECA)~\cite{wang2020eca}, CBAM~\cite{woo2018cbam}, SE-Attention~(seAttn)~\cite{hu2018squeeze}, Depthwise Separable Convolution~(DSC)~\cite{chollet2017xception}, ConvNext~(CN)~\cite{liu2022convnet}. Fig.~\ref{fig:Suppl_Architectures} illustrates the architectural configurations of learnable DR methods, all configured to transform input hypercubes from $C$ spectral channels to $F$ reduced channels ($F \ll C$) while preserving spatial dimensions, $H \times W$. These methods, including LQE, were directly used as the first layer or blocks, whereas for classical DR approaches, a stratified sample selection approach was adopted. Since these are offline sampling-to-projection techniques, we implement a preprocessing pipeline consisting of the following phases on the loaded hypercube and ground truth pairs:

\begin{enumerate}
    \item \textbf{Class-Balanced Pixel Sampling}: A stratified sampling strategy collects 50K representative pixels per dataset. We tally pixel counts per class across all masks and compute per-class sampling targets by dividing the total target samples by the number of classes. For rare classes with fewer pixels than the target, we select all available pixels. For common classes, we randomly sample at calculated rates to achieve class balance. Finally, pixels are sampled from each image proportionally to its class distribution.
    \item \textbf{Normalization}: The sampled pixels are standardized using normalization with statistics computed per spectral band and saved for consistent transformation of all images.
    \item \textbf{Sequential DR}: Each DR method is fitted on the normalized sampled data to learn a transformation to the required components (or, in other terms, filters:~$F$). The learned transformation is then applied to all dataset images through the following steps: (1) flatten each hypercube; (2) normalize using saved statistics; and (3) transform through the fitted model.
\end{enumerate}


\begin{table}[t]
\centering
\caption{Experimental settings and hyperparameters used}
\label{tab:ExperimentalSeteupAndHyperParam}
\resizebox{0.48\textwidth}{!}{%
\begin{tabular}{l c c c}
\toprule
\textbf{Detail} & \textbf{HyKo-VIS} & \textbf{HSI-Drive} & \textbf{H-City}\\
\midrule
Batch ($B$), Accumulate & 16, 1 & 16, 1 & 8, 2\\
\midrule
Optimizer, Seed & \multicolumn{3}{l}{AdamW (0.0001 initial learning rate), and 42} \\
Loss Function & \multicolumn{3}{l}{Combined class-weighted cross entropy and dice loss}\\
Epochs, Scheduler & \multicolumn{3}{l}{300, with early stop of 30 patience on mIoU metric} \\
Hardware & \multicolumn{3}{l}{Intel Xeon Gold, 256GB RAM, Dual Nvidia A6000} \\
Training Time & \multicolumn{3}{l}{Approx. 1 hour for HyKo-VIS to 8 hours of H-City} \\
& \multicolumn{3}{l}{ datasets, per training/ablation configuration.}\\
\bottomrule
\end{tabular}%
}
\end{table}

\textbf{Training Setup}.
All learnable methods are trained end-to-end using mixed precision and early stopping~(patience: 30), without spatial or spectral augmentation. Hyperparameters are detailed in Table~\ref{tab:ExperimentalSeteupAndHyperParam}. To handle class imbalance, a composite loss function $\mathcal{L}_{\text{seg}}$ combining class-weighted cross-entropy and dice losses was used, in the total objective $\mathcal{L}_{\text{total}}$ as defined Eq.~\ref{Eq.12}:

\subsection{Results}
\subsubsection{Quantitative Performance Analysis}
Table~\ref{tab:DR_Wise_Performances} provides comprehensive DR method-wise performance comparisons. LQE demonstrates generalization across the available diverse spectral characteristics datasets: HyKo's VIS spectrum~($C=15$, 470--630~nm), HSI-Drive's extended NIR range~($C=25$, 600--975~nm), and H-City's VIS-to-NIR spectrum~($C=128$, 450--950~nm). Table~\ref{tab:average_performance} presents the overall mean $\pm$ SD values of mIoU, mF1-score, and Kappa across all evaluated SSMs, DR methods, and reduction levels (reducing original dimensions $C$ to $F=3, 5, \text{and }7$).

The proposed LQE achieves consistent improvements across all datasets compared to both conventional and learnable DR methods, respectively: On HyKo-VIS, LQE attains average gains in mIoU (2.45\%, 1.18\%), mF1 (2.40\%, 1.03\%), and Kappa (1.63\%,0.82\%); For HSI-Drive, LQE improves mIoU (0.45\%, 1.56\%), mF1 (0.55\%, 1.63\%), and Kappa (-0.67\%, 1.07\%), while a slight decrease in Kappa is observed relative to conventional DR, LQE consistently outperforms learnable approaches across all metrics; and for H-City, LQE improves mIoU (1.04\%, 0.81\%), mF1 (0.99\%, 0.95\%), and Kappa (0.54\%, 1.16\%).

The individual DR method-wise: Table~\ref{tab:DR_Wise_Performances} and DR category-wise: Table~\ref{tab:average_performance}, results support that LQE provides stable and consistent performance gains across varying spectral ranges, datasets, and input dimensionalities, highlighting its generalizability. Across all configurations and datasets, LQE provides an average improvement of 1.25\%, 1.26\%, and 0.76\% in mIoU, mF1, and Kappa, respectively.

\begin{table*}[htbp]
  \centering
  \caption{Performance of conventional vs. learnable DR methods ($F: 3, 5,$ and $7$) across three datasets: Reported as $\bar{x} \pm SD$, averaged over SSMs. LQE$_P$ results include ablations for $P$. \color{olive} Overall best, \color{teal}best per$F$, and \color{brown}second-best\color{black}. For HSI-Drive and H-City, LLE and Isomap are omitted for low performance.}
  \label{tab:DR_Wise_Performances}
  \vspace{-8pt}
  \textbf{(a) HyKo-VIS}\\
  \resizebox{\textwidth}{!}{%
    \begin{tabular}{p{0.65cm} l *{3}{c}| *{3}{c}| *{3}{c}}
      \toprule
      \multirow{2}{*}{\textbf{Type}} & \textbf{Reduction ($F$)} & \multicolumn{3}{l}{\textbf{Original ($C:$ 15) $\rightarrow$ 3 Channels}} & \multicolumn{3}{c}{\textbf{5 Channels}} & \multicolumn{3}{c}{\textbf{7 Channels}} \\
      \cline{3-5} \cline{6-8} \cline{9-11}
      & \textbf{Methods} & \textbf{mIoU} & \textbf{mF1} & \textbf{Kappa} & \textbf{mIoU} & \textbf{mF1} & \textbf{Kappa} & \textbf{mIoU} & \textbf{mF1} & \textbf{Kappa} \\
      \hline
      \multirow{6}{*}{\rotatebox{90}{\textbf{Conventional}}} & PCA & $58.05 \pm 1.98$ & $71.04 \pm 1.53$ & $81.44 \pm 1.08$ & $58.23 \pm 1.00$ & $71.02 \pm 0.81$ & $82.28 \pm 1.07$ & $57.68 \pm 1.45$ & $70.65 \pm 1.15$ & $82.38 \pm 1.60$ \\
      & ICA & $58.06 \pm 1.67$ & $71.09 \pm 1.13$ & $81.30 \pm 0.89$ & $58.59 \pm 2.30$ & $72.10 \pm 2.36$ & $82.49 \pm 1.72$ & $58.28 \pm 2.20$ & $71.30 \pm 1.96$ & $82.58 \pm 1.42$ \\
      & NMF & $58.87 \pm 2.17$ & $71.68 \pm 1.70$ & $81.60 \pm 1.63$ & $59.23 \pm 1.92$ & $72.04 \pm 1.46$ & $82.19 \pm 1.46$ & $58.75 \pm 1.94$ & $71.72 \pm 1.64$ & \color{olive}$\mathbf{82.66 \pm 1.17}$ \\
      & MNF & $57.94 \pm 2.04$ & $70.92 \pm 1.59$ & $80.89 \pm 0.84$ & $58.14 \pm 2.9$ & $71.26 \pm 2.58$ & $82.53 \pm 1.48$ & $58.08 \pm 1.58$ & $70.87 \pm 1.09$ & $82.52 \pm 1.44$ \\
      & LLE & $50.82 \pm 2.49$ & $64.45 \pm 2.20$ & $74.07 \pm 2.96$ & $44.86 \pm 4.76$ & $58.44 \pm 4.83$ & $68.55 \pm 3.71$ & $47.69 \pm 3.61$ & $61.28 \pm 3.38$ & $73.32 \pm 3.11$ \\
      & Isomap & $57.22 \pm 2.38$ & $70.43 \pm 1.93$ & $80.59 \pm 1.42$ & $56.08 \pm 2.29$ & $68.94 \pm 2.06$ & $80.95 \pm 0.92$ & $57.25 \pm 2.53$ & $70.69 \pm 2.89$ & $80.86 \pm 1.17$ \\
      \cline{2-11}
      & $\bar{\bar{x}} \pm SD_{\bar{x}}$ & $56.83 \pm 2.99$ & $69.94 \pm 2.72$ & $79.98 \pm 2.92$ & $55.85 \pm 5.49$ & $68.97 \pm 5.28$ & $79.83 \pm 5.56$ & $56.29 \pm 4.24$ & $69.42 \pm 4.01$ & $80.72 \pm 3.69$ \\
      \hline
      \multirow{7}{*}{\rotatebox{90}{\textbf{Learnable Layers}}} & 1x1Conv & $56.02 \pm 3.50$ & $69.81 \pm 3.55$ & $79.86 \pm 2.46$ & $57.08 \pm 2.26$ & $70.61 \pm 2.20$ & $80.52 \pm 2.25$ & $57.64 \pm 1.21$ & $70.61 \pm 1.09$ & $80.97 \pm 1.38$ \\
      & AE & $55.58 \pm 1.77$ & $69.47 \pm 1.87$ & $79.04 \pm 1.90$ & $57.38 \pm 2.20$ & $70.50 \pm 1.710$ & $80.72 \pm 1.72$ & $57.23 \pm 1.85$ & $71.34 \pm 2.09$ & $81.01 \pm 1.13$ \\
      & ECA & $59.07 \pm 1.79$ & $71.70 \pm 1.29$ & $81.59 \pm 1.36$ & $58.60 \pm 1.90$ & $71.40 \pm 1.36$ & $82.30 \pm 1.62$ & $58.46 \pm 1.27$ & $71.61 \pm 1.80$ & $81.50 \pm 1.06$ \\
      & CBAM & $56.09 \pm 3.16$ & $69.33 \pm 2.59$ & $79.61 \pm 2.07$ & $57.84 \pm 1.60$ & $70.76 \pm 1.25$ & $81.19 \pm 1.12$ & $58.34 \pm 1.43$ & $71.13 \pm 1.12$ & $81.69 \pm 0.83$ \\
      & seAttn & $57.67 \pm 1.83$ & $70.71 \pm 1.70$ & $81.53 \pm 0.86$ & $58.22 \pm 1.38$ & $71.14 \pm 1.23$ & $81.72 \pm 1.17$ & $58.72 \pm 1.38$ & $72.16 \pm 1.51$ & $81.84 \pm 0.97$ \\
      
      & DSC & $55.29 \pm 2.60$ & $69.15 \pm 3.13$ & $79.24 \pm 2.59$ & $56.62 \pm 2.27$ & $69.94 \pm 1.82$ & $80.28 \pm 1.96$ & $57.82 \pm 1.45$ & $70.79 \pm 1.31$ & $81.10 \pm 1.60$ \\
      & ConvNext & $57.98 \pm 2.60$ & $71.05 \pm 2.27$ & $80.79 \pm 2.22$ & $59.02 \pm 1.93$ & $72.31 \pm 2.05$ & $82.14 \pm 1.60$ & $58.78 \pm 2.40$ & $71.57 \pm 1.74$ & $82.21 \pm 2.2$ \\
      \cline{2-11}
      & $\bar{\bar{x}} \pm SD_{\bar{x}}$ & $56.81 \pm 1.42$ & $70.17 \pm 0.98$ & $80.24 \pm 1.06$ & $57.82 \pm 0.86$ & $70.95 \pm 0.76$ & $81.27 \pm 0.80$ & $58.14 \pm 0.59$ & $71.32 \pm 0.53$ & $81.47 \pm 0.47$\\
      \hline
      \multirow{6}{*}{\rotatebox{90}{\textbf{Proposed}}} & LQE$_1$ & $\mathbf{\color{brown}59.12 \pm 1.95}$ & $\mathbf{\color{brown}71.98 \pm 2.10}$ & $\mathbf{\color{brown}81.80 \pm 1.76}$ & $\mathbf{\color{olive}59.70 \pm 0.88}$ & $\mathbf{\color{olive}72.85 \pm 1.26}$ & $82.26 \pm 1.00$ & $\mathbf{\color{teal}59.37 \pm 1.46}$ & $\mathbf{\color{teal}72.35 \pm 1.52}$ & $82.12 \pm 1.61$ \\
      & LQE$_3$ & $\mathbf{\color{teal}59.49 \pm 1.37}$ & $\mathbf{\color{teal}72.12 \pm 1.12}$ & $\mathbf{\color{teal}82.45 \pm 1.18}$ & $\mathbf{\color{brown}59.31 \pm 0.72}$ & $\mathbf{\color{brown}72.02 \pm 0.56}$ & $\mathbf{\color{brown}82.25 \pm 0.81}$ & $\mathbf{\color{brown}59.31 \pm 1.24}$ & $\mathbf{\color{brown}71.99 \pm 1.11}$ & $\mathbf{\color{brown}82.30 \pm 0.72}$ \\
      & LQE$_5$ & $57.98 \pm 2.21$ & $71.39 \pm 2.48$ & $81.18 \pm 1.98$ & $58.71 \pm 0.64$ & $71.58 \pm 0.63$ & $\mathbf{\color{teal}82.24 \pm 0.62}$ & $58.86 \pm 2.53$ & $71.71 \pm 1.87$ & $81.9 \pm 1.20$ \\
      & LQE$_7$ & $58.06 \pm 1.33$ & $71.17 \pm 1.04$ & $81.40 \pm 1.23$ & $58.48 \pm 1.39$ & $71.31 \pm 1.14$ & $81.70 \pm 1.25$ & $58.20 \pm 1.73$ & $71.18 \pm 1.39$ & $81.73 \pm 1.14$ \\
      & LQE$_9$ & $58.83 \pm 2.35$ & $72.32 \pm 1.66$ & $81.25 \pm 1.71$ & $58.87 \pm 0.67$ & $72.59 \pm 1.66$ & $81.72 \pm 0.96$ & $57.32 \pm 0.68$ & $71.09 \pm 0.49$ & $80.89 \pm 0.72$ \\
      \cline{2-11}
      & $\bar{\bar{x}} \pm SD_{\bar{x}}$ & $58.70 \pm 0.66$ & $71.80 \pm 0.49$ & $81.62 \pm 0.52$ & $59.01 \pm 0.49$ & $72.07 \pm 0.65$ & $82.03 \pm 0.30$ & $58.61 \pm 0.86$ & $71.66 \pm 0.53$ & $81.79 \pm 0.55$ \\
      \bottomrule
    \end{tabular}%
  }
  \\
  \vspace{2.5pt}
  \textbf{(b) HSI-Drive}\\
  \resizebox{\textwidth}{!}{%
    \begin{tabular}{p{0.65cm} l *{3}{c}| *{3}{c}| *{3}{c}}
      \toprule
      \multirow{2}{*}{\textbf{Type}} & \textbf{Reduction ($F$)} & \multicolumn{3}{l}{\textbf{Original ($C:$ 25) $\rightarrow$ 3 Channels}} & \multicolumn{3}{c}{\textbf{5 Channels}} & \multicolumn{3}{c}{\textbf{7 Channels}} \\
      \cline{3-5} \cline{6-8} \cline{9-11}
      & \textbf{Methods} & \textbf{mIoU} & \textbf{mF1} & \textbf{Kappa} & \textbf{mIoU} & \textbf{mF1} & \textbf{Kappa} & \textbf{mIoU} & \textbf{mF1} & \textbf{Kappa} \\
\hline
\multirow{4}{*}{\rotatebox{90}{\textbf{Conv.}}} & PCA & $58.39 \pm 2.75$ & $70.74 \pm 2.85$ & $87.30 \pm 1.13$ & $60.39 \pm 3.05$ & $72.66 \pm 2.96$ & $88.86 \pm 1.44$ & $61.02 \pm 2.17$ & $72.03 \pm 2.19$ & $88.32 \pm 0.91$ \\
& ICA & $57.35 \pm 2.50$ & $70.12 \pm 2.21$ & $87.12 \pm 1.17$ & $59.17 \pm 3.54$ & $71.77 \pm 3.43$ & $88.37 \pm 1.50$ & $60.64 \pm 3.48$ & $72.91 \pm 3.46$ & $88.35 \pm 2.12$ \\
& NMF & $58.49 \pm 2.27$ & $71.20 \pm 2.11$ & $87.41 \pm 1.37$ & $60.33 \pm 3.64$ & $72.60 \pm 3.85$ & $88.25 \pm 1.63$ & $60.56 \pm 1.86$ & $72.63 \pm 1.58$ & $88.31 \pm 0.97$ \\
& MNF & $58.56 \pm 2.82$ & $70.77 \pm 2.81$ & $87.71 \pm 1.57$ & $59.53 \pm 3.05$ & $72.05 \pm 2.85$ & $88.05 \pm 1.55$ & $59.64 \pm 3.07$ & $71.71 \pm 3.09$ & $88.05 \pm 1.47$ \\
\cline{2-11}
      & $\bar{\bar{x}} \pm SD_{\bar{x}}$ & $58.20 \pm 0.57$ & $70.71 \pm 0.44$ & $87.39 \pm 0.25$ & $59.85 \pm 0.60$ & $72.27 \pm 0.43$ & $88.38 \pm 0.34$ & $60.47 \pm 0.59$ & $72.32 \pm 0.55$ & $88.26 \pm 0.14$ \\
\hline
\multirow{7}{*}{\rotatebox{90}{\textbf{Learnable Layers}}} & 1x1Conv & $55.34 \pm 3.41$ & $67.88 \pm 3.34$ & $85.53 \pm 1.55$ & $59.04 \pm 3.69$ & $71.53 \pm 3.35$ & $87.33 \pm 1.72$ & $59.68 \pm 4.48$ & $71.75 \pm 4.21$ & $88.17 \pm 2.33$ \\
& AE & $48.59 \pm 2.33$ & $60.22 \pm 2.50$ & $72.66 \pm 3.55$ & $60.88 \pm 2.18$ & $72.77 \pm 1.83$ & $88.77 \pm 1.28$ & $60.72 \pm 3.37$ & $72.81 \pm 3.29$ & $\mathbf{\color{brown}88.70 \pm 1.57}$ \\
& ECA & $59.20 \pm 3.63$ & $72.45 \pm 3.30$ & $87.83 \pm 2.00$ & $61.02 \pm 2.34$ & $73.46 \pm 3.25$ & $88.53 \pm 1.19$ & $\mathbf{\color{brown}61.15 \pm 2.31}$ & $72.74 \pm 2.20$ & $88.53 \pm 1.29$ \\
& CBAM & $49.11 \pm 2.31$ & $60.20 \pm 2.69$ & $71.98 \pm 3.53$ & $58.25 \pm 3.68$ & $70.69 \pm 3.58$ & $87.14 \pm 2.00$ & $58.11 \pm 4.73$ & $70.54 \pm 4.36$ & $87.11 \pm 2.25$ \\
& seAttn & $59.04 \pm 2.22$ & $72.15 \pm 2.21$ & $87.27 \pm 0.86$ & $60.56 \pm 2.64$ & $73.06 \pm 2.29$ & $88.37 \pm 1.10$ & $60.98 \pm 3.12$ & $72.52 \pm 2.80$ & $88.39 \pm 1.53$ \\
& DSC & $56.36 \pm 4.46$ & $69.28 \pm 4.12$ & $85.84 \pm 2.19$ & $58.62 \pm 3.51$ & $70.89 \pm 3.45$ & $87.37 \pm 1.86$ & $59.36 \pm 2.58$ & $71.51 \pm 2.71$ & $88.07 \pm 1.35$ \\
& ConvNext & $59.26 \pm 4.08$ & $72.45 \pm 3.68$ & $87.20 \pm 2.15$ & $60.77 \pm 3.74$ & $73.08 \pm 3.73$ & $88.52 \pm 2.00$ & $60.27 \pm 1.91$ & $72.53 \pm 1.74$ & $88.33 \pm 1.17$ \\
\cline{2-11}
      & $\bar{\bar{x}} \pm SD_{\bar{x}}$ & $55.27 \pm 4.64$ & $67.80 \pm 5.47$ & $82.62 \pm 7.08$ & $59.88 \pm 1.19$ & $72.21 \pm 1.15$ & $88.00 \pm 0.69$ & $60.04 \pm 1.08$ & $72.06 \pm 0.83$ & $88.19 \pm 0.52$ \\
\hline
\multirow{6}{*}{\rotatebox{90}{\textbf{Proposed}}} & LQE$_1$ & $\mathbf{\color{teal}59.91 \pm 2.27}$ & $\mathbf{\color{teal}72.23 \pm 2.11}$ & $\mathbf{\color{teal}87.88 \pm 1.24}$ & $\mathbf{\color{brown}61.52 \pm 3.31}$ & $\mathbf{\color{brown}73.69 \pm 3.10}$ & $\mathbf{\color{brown}88.91 \pm 1.99}$ & $\mathbf{\color{teal}61.20 \pm 2.74}$ & $\mathbf{\color{teal}73.32 \pm 2.63}$ & $88.32 \pm 1.60$ \\
& LQE$_3$ & $\mathbf{\color{brown}59.92 \pm 3.52}$ & $\mathbf{\color{brown}72.41 \pm 3.23}$ & $\mathbf{\color{brown}87.83 \pm 1.65}$ & $\mathbf{\color{olive}61.86 \pm 4.04}$ & $\mathbf{\color{olive}73.96 \pm 3.73}$ & $\mathbf{\color{olive}88.96 \pm 1.81}$ & $60.27 \pm 3.90$ & $72.52 \pm 3.62$ & $88.11 \pm 1.90$ \\
& LQE$_5$ & $58.96 \pm 1.97$ & $71.32 \pm 1.99$ & $87.52 \pm 1.18$ & $59.12 \pm 3.31$ & $71.45 \pm 3.06$ & $87.36 \pm 1.40$ & $60.14 \pm 3.21$ & $72.32 \pm 2.94$ & $88.14 \pm 1.41$ \\
& LQE$_7$ & $59.38 \pm 4.29$ & $71.69 \pm 4.04$ & $87.49 \pm 2.23$ & $59.57 \pm 2.24$ & $71.84 \pm 1.94$ & $87.84 \pm 1.27$ & $61.02 \pm 3.36$ & $\mathbf{\color{brown}73.36 \pm 3.11}$ & $\mathbf{\color{teal}88.64 \pm 1.06}$ \\
& LQE$_9$ & $59.15 \pm 2.35$ & $71.83 \pm 2.05$ & $84.38 \pm 3.56$ & $59.29 \pm 2.91$ & $72.07 \pm 2.96$ & $84.72 \pm 3.44$ & $58.14 \pm 2.89$ & $70.84 \pm 2.62$ & $83.99 \pm 3.53$ \\
\cline{2-11}
& $\bar{\bar{x}} \pm SD_{\bar{x}}$ & $59.46 \pm 0.44$ & $71.90 \pm 0.43$ & $87.02 \pm 1.49$ & $60.27 \pm 1.31$ & $72.60 \pm 1.14$ & $87.56 \pm 1.73$ & $60.15 \pm 1.22$ & $72.47 \pm 1.02$ & $87.44 \pm 1.94$ \\
\bottomrule
\end{tabular}%
}

  \vspace{2.5pt}
  \textbf{(c) H-City}\\
  \resizebox{\textwidth}{!}{%
    \begin{tabular}{p{0.65cm} l *{3}{c}| *{3}{c}| *{3}{c}}
      \toprule
      \multirow{2}{*}{\textbf{Type}} & \textbf{Reduction ($F$)} & \multicolumn{3}{l}{\textbf{Original ($C:$ 128) $\rightarrow$ 3 Channels}} & \multicolumn{3}{c}{\textbf{5 Channels}} & \multicolumn{3}{c}{\textbf{7 Channels}} \\
      \cline{3-5} \cline{6-8} \cline{9-11}
      & \textbf{Methods} & \textbf{mIoU} & \textbf{mF1} & \textbf{Kappa} & \textbf{mIoU} & \textbf{mF1} & \textbf{Kappa} & \textbf{mIoU} & \textbf{mF1} & \textbf{Kappa} \\
\hline
\multirow{4}{*}{\rotatebox{90}{\textbf{Conv.}}} & PCA & $49.08 \pm 2.96$ & $61.10 \pm 3.03$ & $88.12 \pm 1.83$ & $49.83 \pm 2.07$ & $61.02 \pm 1.95$ & $87.79 \pm 1.76$ & $49.87 \pm 2.02$ & $61.75 \pm 1.99$ & $88.27 \pm 1.47$ \\
& ICA & $48.83 \pm 1.87$ & $60.82 \pm 1.69$ & $87.39 \pm 1.61$ & $47.94 \pm 3.69$ & $60.30 \pm 3.22$ & $86.34 \pm 2.61$ & $48.89 \pm 2.47$ & $60.87 \pm 2.16$ & $87.13 \pm 2.08$ \\
& NMF & $48.82 \pm 2.49$ & $61.44 \pm 2.08$ & $88.55 \pm 1.13$ & $50.34 \pm 2.34$ & $61.73 \pm 2.30$ & $89.06 \pm 1.33$ & $50.28 \pm 2.00$ & $62.93 \pm 1.47$ & $88.73 \pm 1.61$ \\
& MNF & $48.72 \pm 2.64$ & $60.59 \pm 2.72$ & $87.98 \pm 1.75$ & $50.28 \pm 1.99$ & $61.40 \pm 1.53$ & $88.24 \pm 1.53$ & $50.11 \pm 1.77$ & $62.15 \pm 1.54$ & $88.00 \pm 1.64$ \\
\cline{2-11}
      & $\bar{\bar{x}} \pm SD_{\bar{x}}$ & $48.86 \pm 0.15$ & $60.99 \pm 0.37$ & $88.01 \pm 0.48$ & $49.60 \pm 1.13$ & $61.11 \pm 0.61$ & $87.86 \pm 1.14$ & $49.79 \pm 0.62$ & $61.93 \pm 0.86$ & $88.03 \pm 0.67$ \\
\hline
\multirow{7}{*}{\rotatebox{90}{\textbf{Learnable Layers}}} & 1x1Conv & $49.05 \pm 0.46$ & $61.15 \pm 1.19$ & $87.27 \pm 0.04$ & $49.92 \pm 1.75$ & $61.90 \pm 1.73$ & $88.16 \pm 1.20$ & $51.69 \pm 2.60$ & $63.46 \pm 3.20$ & $88.85 \pm 1.57$ \\
& AE & $48.73 \pm 6.69$ & $60.84 \pm 6.49$ & $86.94 \pm 4.06$ & $50.71 \pm 3.27$ & $62.14 \pm 4.19$ & $88.25 \pm 2.80$ & $\mathbf{\color{brown}52.08 \pm 3.68}$ & $64.24 \pm 3.89$ & $\mathbf{\color{teal}89.80 \pm 1.55}$ \\
& ECA & $49.50 \pm 1.54$ & $61.07 \pm 1.55$ & $\mathbf{\color{brown}88.62 \pm 1.16}$ & $50.29 \pm 3.26$ & $62.25 \pm 1.81$ & $88.48 \pm 1.55$ & $51.28 \pm 2.05$ & $63.06 \pm 2.14$ & $88.75 \pm 1.87$ \\
& CBAM & $46.92 \pm 6.49$ & $58.99 \pm 6.50$ & $85.65 \pm 3.65$ & $\mathbf{\color{brown}51.68 \pm 4.81}$ & $63.04 \pm 4.09$ & $\mathbf{\color{teal}89.14 \pm 1.24}$ & $51.43 \pm 5.78$ & $62.96 \pm 5.06$ & $88.50 \pm 2.71$ \\
& seAttn & $49.49 \pm 2.82$ & $61.23 \pm 2.96$ & $88.61 \pm 1.67$ & $51.01 \pm 2.84$ & $62.41 \pm 2.91$ & $88.31 \pm 1.69$ & $51.68 \pm 2.88$ & $63.37 \pm 2.66$ & $89.21 \pm 1.75$ \\
& DSC & $40.22 \pm 2.98$ & $50.54 \pm 2.37$ & $74.46 \pm 2.91$ & $44.63 \pm 3.05$ & $56.05 \pm 3.34$ & $80.18 \pm 2.42$ & $49.66 \pm 4.12$ & $61.90 \pm 4.11$ & $87.90 \pm 2.13$ \\
& ConvNext & $\mathbf{\color{brown}49.62 \pm 2.15}$ & $61.45 \pm 1.98$ & $\mathbf{\color{teal}88.70 \pm 1.07}$ & $51.52 \pm 3.26$ & $63.14 \pm 3.10$ & $\mathbf{\color{brown}89.13 \pm 2.09}$ & $51.59 \pm 3.18$ & $63.74 \pm 3.12$ & $\mathbf{\color{brown}89.36 \pm 1.75}$ \\
\cline{2-11}
      & $\bar{\bar{x}} \pm SD_{\bar{x}}$ & $47.65 \pm 3.40$ & $59.32 \pm 3.96$ & $85.75 \pm 5.10$ & $49.97 \pm 2.44$ & $61.56 \pm 2.47$ & $87.38 \pm 3.20$ & $51.34 \pm 0.78$ & $63.25 \pm 0.73$ & $88.91 \pm 0.62 $ \\
\hline
\multirow{5}{*}{\rotatebox{90}{\textbf{Proposed}}} & LQE$_1$ &  $49.47 \pm 2.64$ & $61.64 \pm 2.74$ & $88.27 \pm 1.98$ & $50.08 \pm 3.54$ & $61.63 \pm 3.62$ & $88.70 \pm 1.84$ & $51.99 \pm 1.95$ & $\mathbf{\color{brown}63.80 \pm 2.18}$ & $89.13 \pm 1.67$ \\
& LQE$_3$ & $\mathbf{\color{teal}49.63 \pm 2.43}$ & $\mathbf{\color{brown}61.58 \pm 2.37}$ & $88.29 \pm 2.10$ & $\mathbf{\color{teal}52.06 \pm 2.62}$ & $\mathbf{\color{teal}63.58 \pm 2.35}$ & $89.01 \pm 1.55$ & $50.20 \pm 3.61$ & $61.87 \pm 3.41$ & $88.53 \pm 1.96$ \\
& LQE$_5$ & $\mathbf{\color{brown}49.62 \pm 2.19}$ & $\mathbf{\color{teal}61.96 \pm 2.42}$ & $88.17 \pm 1.37$ & $49.31 \pm 2.02$ & $61.45 \pm 2.49$ & $87.85 \pm 1.40$ & $\mathbf{\color{olive}53.09 \pm 2.07}$ & $\mathbf{\color{olive}64.85 \pm 2.42}$ & $88.48 \pm 1.68$ \\
& LQE$_7$ & $49.22 \pm 4.26$ & $60.75 \pm 4.24$ & $88.13 \pm 1.92$ & $50.87 \pm 2.87$ & $\mathbf{\color{brown}63.16 \pm 2.71}$ & $88.79 \pm 1.71$ & $50.02 \pm 1.85$ & $61.70 \pm 1.88$ & $88.78 \pm 1.04$ \\
\cline{2-11}
      & $\bar{\bar{x}} \pm SD_{\bar{x}}$ & $49.48 \pm 0.19$ & $61.48 \pm 0.52$ & $88.22 \pm 0.08$ & $50.58 \pm 1.17$ & $62.46 \pm 1.07$ & $88.59 \pm 0.51$ & $51.33 \pm 1.47$ & $63.05 \pm 1.53$ & $88.73 \pm 0.30$ \\
\bottomrule
\multicolumn{11}{l}{\textsuperscript{*} $\bar{x}$: Mean. \quad $SD$: Standard Deviation. \quad $\bar{\bar{x}}$: Method-wise Mean. \quad $SD_{\bar{x}}$: Standard Deviation over method-wise $\bar{x}$}\\
\multicolumn{11}{l}{$^\dagger$ LQE$_9$ excluded from H-City (C\,=\,128, 1422$\times$1889), due to consistent diminishing returns for $P \geq 5$ on other datasets.}
\end{tabular}%
}
\end{table*}

\begin{table*}[t]
  \centering
  \caption{Overall average performance across categorical DR methods, over $F=3,5,7$ and SSMs: Individual method-wise details are shown in Table~\ref{tab:DR_Wise_Performances}.}
  \label{tab:average_performance}
  \resizebox{0.99\textwidth}{!}{
    \begin{tabular}{l ccc ccc ccc}
      \toprule
      \multirow{2}{*}{\rotatebox{0}{\textbf{Category}}} & \multicolumn{3}{c}{\textbf{HyKo-VIS}} & \multicolumn{3}{c}{\textbf{HSI-Drive}} & \multicolumn{3}{c}{\textbf{H-City}} \\
      \cmidrule(lr){2-4} \cmidrule(lr){5-7} \cmidrule(lr){8-10}
      & \textbf{mIoU} & \textbf{mF1} & \textbf{Kappa} & \textbf{mIoU} & \textbf{mF1} & \textbf{Kappa} & \textbf{mIoU} & \textbf{mF1} & \textbf{Kappa}\\
      \midrule
      \textbf{Orig.}~\textsuperscript{a} & $64.48 \pm 3.56$ &$76.29 \pm 3.27$ & $90.32 \pm 1.74$ & $60.51 \pm 0.87$ & $73.47 \pm 1.13$ & $83.08 \pm 0.88$ & $53.30 \pm 2.28$ & $65.23 \pm 2.24$ & $89.99 \pm 1.18$\\
      \midrule
      \textbf{Conv.}~\textsuperscript{b} & $56.32 \pm 4.12$ & $69.44 \pm 3.91$ & $80.18 \pm 3.97$ & $59.51 \pm 1.13$ & $71.77 \pm 0.89$ & $88.01 \pm 0.52$ & $49.42 \pm 0.80$ & $61.34 \pm 0.73$ & $87.97 \pm 0.74$\\
      \textbf{Learn.}~\textsuperscript{c} & $57.59 \pm 1.13$ & $70.81 \pm 0.89$ & $80.99 \pm 0.95$ & $58.40 \pm 3.52$ & $70.69 \pm 3.74$ & $86.27 \pm 4.72$ & $49.65 \pm 2.81$ & $61.38 \pm 3.07$ & $87.35 \pm 3.57$\\
      \textbf{LQE}  & $58.77 \pm 0.66$ & $71.84 \pm 0.55$ & $81.81 \pm 0.47$ & $59.96 \pm 1.05$ & $72.32 \pm 0.91$ & $87.34 \pm 1.62$ & $50.46 \pm 1.27$ & $62.33 \pm 1.22$ & $88.51 \pm 0.38$\\
      
      \bottomrule
      \multicolumn{7}{l}{\textsuperscript{a} Original Dataset Channels (Table~\ref{tab:hsi_datasets}).\quad \textsuperscript{b} Conventional. \quad \textsuperscript{c} Learnable.}\\
    \end{tabular}}%
\end{table*}

\begin{sidewaystable}
    \centering
    \caption{Class-wise IoU results on the HyKo-VIS dataset for the top--10 mIoU-wise DR methods (sorted from left to right), reduced to $F=3$, and six SSMs operating on the original dataset $C$ channels. Performance highlights indicate the \textbf{\color{olive}best} and \textbf{\color{teal}second-best} performing technique for each respective class.}
    \label{Suppl_tab:HyKo_classComparison}
    \resizebox{\textwidth}{!}{%
        \begin{tabular}{ l *{10}{>{\centering\arraybackslash}p{0.066\textwidth}} | *{6}{>{\centering\arraybackslash}p{0.066\textwidth}} }
            \toprule
            & \multicolumn{10}{c}{$F=3$} & \multicolumn6{c}{\textbf{Original Dataset ($C=15$)}}\\
            \cline{2-17}
            \textbf{DR Method$\rightarrow$} & \textbf{LQE$_1$} & \textbf{LQE$_3$} & \textbf{LQE$_9$} & \textbf{ECA} & \textbf{PCA} & \textbf{NMF} & \textbf{ECA} & \textbf{MNF} & \textbf{LQE$_9$} & \textbf{LQE$_5$} &  \\
            \textbf{SSMs $\rightarrow$} & \textbf{UNet++} & \textbf{UNet++} & \textbf{DeepLabV3+} & \textbf{PSPNet} & \textbf{UNet++} & \textbf{UNet++} & \textbf{UNet++} & \textbf{UNet++} & \textbf{UNet++} & \textbf{UNet++} & \textbf{PSPNet} & \textbf{DeepLabV3+} & \textbf{UNet++} & \textbf{UNet} & \textbf{Segformer} & \textbf{FPN} \\
            \hline
            Road & 85.81 & \textbf{\color{olive}86.81} & 85.25 & 83.13 & 83.89 & 84.18 & 85.45 & 83.58 & 84.28 & \textbf{\color{teal}86.39} & 85.26 & 84.25 & 83.94 & 85.67 & 85.15 & 82.80 \\
            Sidewalk & 71.39 & 76.15 & 74.67 & 73.07 & 73.57 & \textbf{\color{olive}80.38} & 69.74 & 71.37 & 68.38 & 69.87 & 69.74 & 73.33 & 62.72 & 71.34 & 79.51 & 67.68 \\
            Lane & 53.57 & 51.12 & \textbf{\color{olive}54.59} & 45.42 & 52.01 & 52.42 & 51.65 & 50.75 & 53.11 & \textbf{\color{teal}54.18} & 47.66 & 46.85 & 45.53 & 49.97 & 47.35 & 51.93 \\
            Grass & \textbf{\color{teal}69.39} & 65.67 & 63.96 & 72.25 & 60.42 & 63.24 & 68.45 & 65.03 & 60.87 & 61.84 & \textbf{\color{olive}73.68} & 69.19 & 63.06 & 62.56 & 60.78 & 66.28 \\
            Vegetation & 62.82 & 60.77 & 58.38 & 60.49 & 63.50 & \textbf{\color{olive}64.33} & 62.56 & 57.89 & 58.79 & 60.69 & \textbf{\color{teal}63.98} & 57.44 & 60.20 & 58.72 & 61.39 & 59.99 \\
            Panels\textsuperscript{*} & 52.16 & 51.96 & \textbf{\color{olive}57.19} & \textbf{\color{teal}52.82} & 50.46 & 51.30 & 47.49 & 52.59 & 49.49 & 47.54 & 48.47 & 52.78 & 48.86 & 51.46 & 51.51 & 48.77 \\
            Buildings & \textbf{\color{olive}53.39} & 53.06 & 51.23 & 48.76 & 48.09 & 47.00 & 52.53 & 50.20 & 50.65 & 49.90 & \textbf{\color{teal}53.70} & 50.36 & 51.88 & 44.53 & 47.69 & 49.68 \\
            Car & 21.93 & 22.06 & 14.68 & 24.84 & \textbf{\color{teal}26.66} & 19.77 & 21.80 & 26.53 & 25.97 & 21.44 & \textbf{\color{olive}26.67} & 19.32 & 26.25 & 19.66 & 13.19 & 22.78 \\
            Person & 27.55 & 22.43 & 26.94 & 26.51 & 25.25 & 20.64 & 23.29 & 24.25 & \textbf{\color{teal}28.92} & 27.44 & 25.47 & 24.86 & \textbf{\color{olive}35.38} & 26.50 & 22.21 & 19.60 \\
            Sky & \textbf{\color{olive}97.51} & 96.82 & 95.23 & 96.76 & 97.06 & 96.47 & 96.93 & 97.01 & 96.56 & 96.37 & 96.93 & 96.95 & 96.46 & 97.00 & \textbf{\color{teal}97.46} & 96.71 \\
            \hline
            mIoU          & \textbf{\color{olive}62.79} & \textbf{\color{teal}61.94} & 61.57 & 61.55 & 61.35 & 61.28 & 61.24 & 61.12 & 60.88 & 60.75 & 62.22 & 60.73 & 60.59 & 59.91 & 59.89 & 59.73 \\
            mF1           & \textbf{\color{olive}75.97} & 74.14 & 73.72 & 73.65 & 73.49 & 73.30 & 73.33 & 73.33 & 73.17 & 75.46& 74.30 & 73.32 & \textbf{\color{teal}75.56} & 72.41 & 72.75 & 72.50 \\
            mAccuracy     & \textbf{\color{olive}97.45} & \textbf{\color{teal}97.34} & 97.11 & 97.13 & 97.01 & 97.08 & 97.22 & 96.94 & 97.01 & 97.20 & 97.42 & 97.05 & 97.03 & 96.99 & 97.00 & 97.05 \\
            mSpecificity  & \textbf{\color{olive}98.42} & \textbf{\color{teal}98.38} & 98.20 & 98.19 & 98.14 & 98.21 & 98.30 & 98.09 & 98.15 & 98.30 & 98.37 & 98.16 & 98.20 & 98.18 & 98.14 & 98.15 \\
            Kappa         & \textbf{\color{olive}85.18} & 84.61 & 83.19 & 83.33 & 82.59 & 82.97 & 83.90 & 82.15 & 82.64 & 83.77 & \textbf{\color{teal}84.92} & 82.83 & 82.81 & 82.55 & 82.51 & 82.83 \\
            \bottomrule
            \multicolumn{17}{l}{\textsuperscript{*} Traffic Sign/Boards.}\\
        \end{tabular}%
    }
\end{sidewaystable}

\begin{sidewaystable}
    \centering
    \caption{Class-wise IoU results on the HSI-Drive Datasets for the top--10 mIoU-wise DR methods (sorted from left to right), reduced to $F=3$, and six SSMs operating on the original dataset $C$ channels. Performance highlights indicate the \textbf{\color{olive}best} and \textbf{\color{teal}second-best} performing technique for each respective class.}
    \label{Suppl_tab:HSIDrive_classComparison}
    \resizebox{\textwidth}{!}{%
        \begin{tabular}{ l *{10}{>{\centering\arraybackslash}p{0.066\textwidth}} | *{6}{>{\centering\arraybackslash}p{0.066\textwidth}} }
            \toprule
            & \multicolumn{10}{c}{$F=3$} & \multicolumn6{c}{\textbf{Original Dataset ($C=25$)}}\\
            \cline{2-17}
            \textbf{DR Method$\rightarrow$} & \textbf{ECA} & \textbf{LQE$_7$} & \textbf{LQE$_3$} & \textbf{LQE$_1$} & \textbf{LQE$_3$} & \textbf{CBAM} & \textbf{NMF} & \textbf{seAttn} & \textbf{LQE$_7$} & \textbf{seAttn} & \\

            \textbf{SSMs $\rightarrow$} & \textbf{DeepLabV3$+$} & \textbf{UNet$++$} & \textbf{UNet$++$} & \textbf{UNet$++$} & \textbf{DeepLabV3$+$} & \textbf{UNet$++$} & \textbf{DeepLabV3$+$} & \textbf{UNet$++$} & \textbf{DeepLabV3$+$} & \textbf{PSPNet} & \textbf{DeepLabV3$+$} & \textbf{UNet$++$} & \textbf{PSPNet} & \textbf{Segformer} & \textbf{FPN} & \textbf{UNet} \\
            \hline
            
            Road & \textbf{97.40} & 97.33 &  96.94 & 96.98 & 97.27 & 97.03 & 97.09 & 96.64 & 97.36 & 96.84 & \textbf{\color{olive}97.59} & \textbf{\color{teal}97.51} & 97.02 & 97.08 & 96.95 & 95.86 \\
            Road Marks & 83.39 & \textbf{\color{teal}85.22} & 83.40 & 83.45 & 83.51 & 84.38 & 80.96 & 80.81 & \textbf{\color{olive}85.35} & 76.19 & 84.87 & 83.15 & 75.08 & 77.14 & 79.79 & 78.25 \\
            Vegetation & \textbf{91.78} & 89.21 & 87.39 & 87.77 & 85.20 & 85.14 & 88.78 & 88.87 & 84.26 & 90.18 & 91.94 & 91.11 & \textbf{\color{olive}92.96} & \textbf{\color{teal}91.99} & 90.23 & 87.03 \\
            P.Mtl\textsuperscript{a} & 39.93 & \textbf{43.02} & 33.67 & 37.49 & 34.65 & 40.13 & 38.47 & 38.71 & 33.97 & 36.62& \textbf{\color{teal}46.14} & \textbf{\color{olive}47.27} & 43.50 & 39.77 & 36.25 & 34.74 \\
            Sky & 89.92 & 88.28 & 88.33 & 89.59 & 87.16 & 84.72 & \textbf{90.61} & 89.86 & 86.88 & 89.38& 89.82 & 90.04 & \textbf{\color{teal}91.17} & \textbf{\color{olive}92.61} & 88.86 & 84.12 \\
            Conc.\textsuperscript{b} & \textbf{58.40} & 53.85 & 46.80 & 52.98 & 50.03 & 46.72 & 50.67 & 47.01 & 47.09 & 56.95 & \textbf{\color{teal}58.90} & 57.23 & \textbf{\color{olive}61.04} & 55.20 & 52.92 & 40.84 \\
            Ped\textsuperscript{c} & 27.08 & 27.42 & \textbf{34.24} & 20.19 & 25.11 & 25.01 & 17.14 & 20.98 & 19.31 & 24.71 & \textbf{\color{teal}36.07} & 25.85 & \textbf{\color{olive}39.71} & 31.08 & 20.71 & 17.53 \\
            U.Mtl\textsuperscript{d} & 50.96 & 50.51 & 50.39 & 49.42 & 51.62 & 50.31 & 48.46 & 47.17 & \textbf{\color{teal}53.25} & 48.03 & 52.83 & \textbf{\color{olive}53.82} & 50.50 & 49.31 & 43.92 & 38.35 \\
            Glass\textsuperscript{e} & 46.00 & \textbf{\color{teal}49.53}  & 47.62 & 47.20 & 47.57 & 47.24 & 45.29 & 47.50 & 49.50 & 38.09 & \textbf{\color{olive}49.75} & 48.84 & 40.62 & 40.77 & 43.21 & 41.39 \\
            \hline
            mIoU & \textbf{65.77} & 65.67 & 63.99 & 63.56 & 63.25 & 63.08 & 62.71 & 62.71 & 62.65 & 62.63 & \textbf{\color{olive}68.40} & \textbf{\color{teal}66.93} & 66.44 & 64.68 & 62.19 & 58.25 \\
            mF1 & 77.55 & \textbf{77.63} & 75.76 & 75.56 & 75.65 & 75.48 & 75.27 & 74.27 & 75.09 & 74.68 & \textbf{\color{olive}79.70} & \textbf{\color{teal}78.34} & 78.23 & 76.57 & 74.43 & 70.49 \\
            mAccuracy & \textbf{98.88} & 98.78 & 98.57 & 98.66 & 98.63 & 98.54 & 98.71 & 98.58 & 98.58 & 98.74 & \textbf{\color{olive}98.94} & \textbf{\color{teal}98.91} & 98.90 & 98.84 & 98.73 & 98.34 \\
            mSpecificity & \textbf{99.23} & 99.18 & 99.09 & 99.15 & 99.04 & 99.02 & 99.10 & 99.07 & 99.03 & 99.13 & \textbf{\color{olive}99.32} & \textbf{\color{teal}99.27} & 99.26 & 99.18 & 99.13 & 98.92 \\
            Kappa & \textbf{91.16} & 90.35 & 88.64 & 89.39 & 88.99 & 88.27 & 89.73 & 88.66 & 88.67 & 90.05 & \textbf{\color{olive}91.62} & 91.26 & \textbf{\color{teal}91.52} & 90.76 & 89.88 & 86.85 \\
            \bottomrule
            \multicolumn{17}{l}{\textsuperscript{a} Painted Metal. \quad\textsuperscript{b} Concrete/Stone/Brick.    \quad\textsuperscript{c} Pedestrian/Cyclist.    \quad\textsuperscript{d} Unpainted metal.  \quad\textsuperscript{e} Glass/Transparent plastic.}
        \end{tabular}%
    }

\end{sidewaystable}

\begin{figure}[t]
  \centering
   \includegraphics[width=\columnwidth]{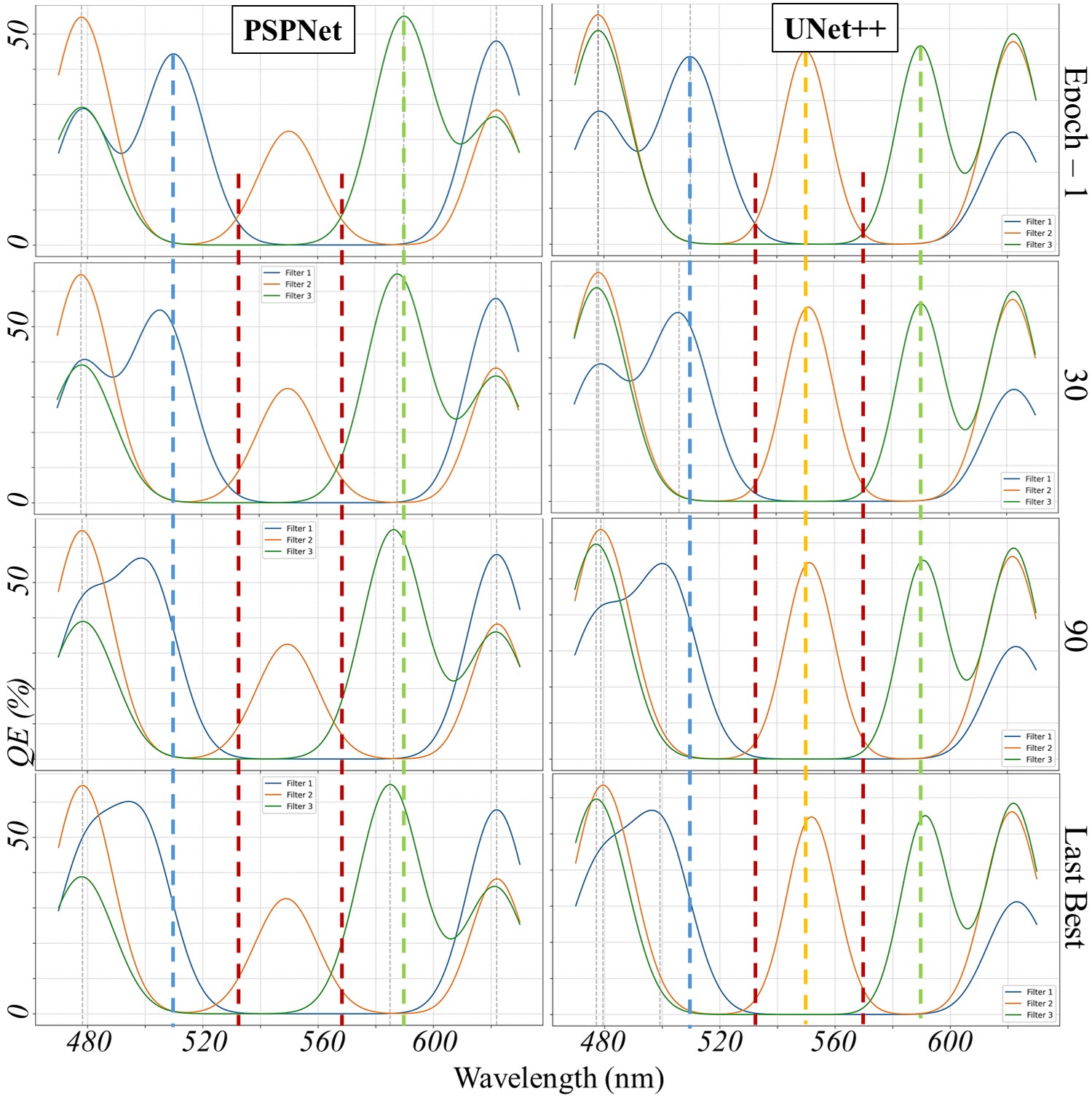}
   \caption{LQE spectral response filters on HyKo-VIS dataset with $F=3$ and $P=3$. Filters are trained on different backbones and initialized with random perturbations as in Epoch-1. Both architectures converge to similar spectral response patterns by the final epoch, as evidenced by the substantial shift in Filter 1 (blue curves) from the blue-green region ($\sim$510 nm) toward the blue region ($\sim$490 nm). Solid curves represent the complete LQE filter responses, while dashed vertical lines indicate starting peak centroids (blue/green). Red dashed lines highlight spectral regions where filter coverage transitions, showing how two adjacent filters adapt to maintain complete spectral coverage as one filter shifts, specifically at $\sim$530nm.}
   \label{fig:lqe_adaptability_SSMS}
\end{figure}

\subsection{Ablation Studies}
\subsubsection{Impact of Peak Count}
We systematically evaluate the effect of $P$ for each output $F \in \{3, 5, 7\}$ dimension to understand the trade-off between model expressiveness and interpretability. Table~\ref{tab:DR_Wise_Performances} shows that performance generally improves with increasing $P$ up to~3. On average, $P\ge5$ provides diminishing returns, while parameter count increases linearly. Based on this analysis, $P=3$ provides the optimal trade-off across datasets. However, for the wider-band H-City dataset, $P=5$ yields gains at $F=7$, suggesting that the dataset's spectral complexity influences the optimal $P$ count.

\subsubsection{Spectral Response Filter Visualization}
Fig.~\ref{fig:lqe_adaptability_SSMS} demonstrates LQE's training on HyKo-VIS. Despite different initializations and architectures (PSPNet vs. UNet++), filters converge to consistent spectral patterns, with Filter 1 (blue solid line) migrating from $\sim$510 nm to $\sim$490 nm, and adjacent filter overlap regions shift to maintain full spectral coverage. This architecture-agnostic convergence validates that LQE is capable of discovering dataset-intrinsic spectral response filters.


\subsubsection{Class-Wise Performance Analysis}
\label{subsec:ClassWisePerformance}
Tables~\ref{Suppl_tab:HyKo_classComparison}--\ref{Suppl_tab:HSIDrive_classComparison} present class-wise IoU results for the HyKo-VIS and HSI-Drive datasets, respectively. These tables compare the top--10 performing DR methods (sorted by mIoU), each reducing the original HyKo-VIS~($C=15$) and HSI-Drive~($C=25$) dimensions to $F=3$. The tables also show performance on the original dataset $C$ channels, across all six SSMs. Following are the key observations and findings from these results:
\begin{itemize}
    \item For HyKo-VIS, LQE$_1$ (LQE$_{P=1}$) configuration with UNet++ achieves the highest overall mIoU (62.79\%) and mF1 (75.97\%), followed by the LQE$_3$ on UNet++ with mIoU of 61.94\%. Notably, the proposed LQE outperforms even the models trained on the original $C=15$ channels. While the best performance for each class depends on the specific DR method and SSM combination, the LQE configurations demonstrate strong performance across critical classes: Road (86.81\%), Lanes (54.59\%), and Person (28.92\%) in best-case results. LQE configurations occupy multiple top-10 positions, demonstrating robustness across configurations.
    \item For HSI-Drive, models trained on the original $C=25$ channels performed better, whereas LQE variants achieve competitive performance compared to the full-dimensional inputs. However, LQE excels particularly in critical classes such as road markings, pedestrians, painted metal, and unpainted metal, consistently ranking among the top performers (best, second-best, or top-10) among all learnable DR methods. ECA with DeepLabV3+ achieves the highest mIoU (65.77\%), closely followed by LQE$_7$ (65.67\%). For HSI-Drive, LQE demonstrates particular strength in road markings (85.35\%), pedestrians (34.24\%), glass/transparent plastic (49.53\%), painted metal (43.02\%), and unpainted metal (53.25\%) accuracies.
\end{itemize}

\begin{table}[h]
\centering
\caption{Regularization components: HyKo-VIS over UNet++.}
\label{tab:reg_ablation}
\begin{tabular}{lcc|cc}
\toprule
\multirow{2}{*}{\textbf{Configuration}} & \multicolumn{2}{c}{\textbf{($F=3$, $P=1$)}} & \multicolumn{2}{c}{\textbf{($F=3$, $P=3$)}}\\
\cline{2-5}
& \textbf{mIoU} & \textbf{mF1} & \textbf{mIoU} & \textbf{mF1}\\
\hline
with $\mathcal{L}_{\text{sep}}$ & 59.54 & 71.73 & 58.99 & 72.43\\
with $\mathcal{L}_{\text{bw}}$ & 60.61 & 72.44 & 60.13 & 73.44\\
with $\mathcal{L}_{\text{dom}}$ & 61.21 & 73.29 & 61.25 & 73.25\\
\hline
Full LQE ($\mathcal{L}_{\text{reg}}$) & \textbf{62.79} & \textbf{75.97} & \textbf{61.94} & \textbf{74.14}\\
\bottomrule
\end{tabular}
\end{table}

\subsubsection{Impact of Regularization Components}
The regularization components in this work are motivated by physical CFA QE curve characteristics. Specifically, single dominant peaks, bounded bandwidth, and centroid separability, as shown in Fig.~\ref{fig:lqe_Inspiration_Figure}. These constraints were originally intended to ensure physical realism and inform multispectral system design rather than to improve quantitative performance. However, they provide improvements in overall performance while simultaneously ensuring physical interpretability and plausibility, as demonstrated in Fig.~\ref{fig:lqe_adaptability_SSMS}. Table~\ref{tab:reg_ablation} evaluates the contribution of individual regularization terms by introducing each component to $\mathcal{L}_{\text{seg}}$. $\mathcal{L}_{\text{dom}}$ contributes most significantly, followed by $\mathcal{L}_{\text{bw}}$ and $\mathcal{L}_{\text{sep}}$. 

\subsubsection{Architecture-Agnostic Spectral Learning}

Tables~\ref{Suppl_tab:HyKo_classComparison}--\ref{Suppl_tab:HSIDrive_classComparison} show that LQE variants integrated with SSMs achieve competitive performance across backbones, with UNet++ and DeepLabV3+ consistently among the top performers. This suggests that the learned spectral response filters integrate effectively with diverse architectures while capturing dataset-specific patterns. Despite different initialisations and distinct SSM backbones (PSPNet vs.\ UNet++) on the HyKo-VIS dataset, as shown in Fig.~\ref{fig:lqe_adaptability_SSMS}, the LQE filters converge to similar spectral responses, indicating that they primarily reflect dataset characteristics rather than architecture-specific representations.

\begin{table}[h]
\caption{Computational Performance ($F=3$), not accounting for downstream SSMs performance.}
\resizebox{\columnwidth}{!}{
\centering
\begin{tabular}{l l r r r}
\toprule
\textbf{Dataset} & \textbf{Method} & \textbf{Param (Million)} & \textbf{CPU (ms)~\textsuperscript{*  }} & \textbf{GPU (ms)~\textsuperscript{*  }} \\
\midrule
\multirow{10}{*}{\textbf{HyKo-VIS}} & 1x1Conv & 51 & $2.92 \pm 0.89$ & $\mathbf{0.16 \pm 0.05}$ \\
& AE & 846 & $2.59 \pm 0.27$ & $0.40 \pm 0.02$ \\
& ECA & 52 & $1.44 \pm 0.10$ & $0.29 \pm 0.08$ \\
& CBAM & 155 & $3.88 \pm 0.22$ & $5.70 \pm 2.09$ \\
& seAttn & 63 & $\mathbf{1.08 \pm 0.06}$ & $0.28 \pm 0.04$ \\
& DSC & 204 & $2.71 \pm 0.34$ & $0.25 \pm 0.38$ \\
& ConvNeXt & 1K & $3.96 \pm 0.25$ & $0.28 \pm 0.01$ \\
\cmidrule{2-5}
& LQE$_1$ & \textbf{12} & $3.03 \pm 0.83$ & $ 1.30\pm 0.09$ \\
& LQE$_3$ & \textbf{36} & $3.24 \pm 0.56$ & $1.73 \pm 0.20$ \\
\midrule
\multirow{10}{*}{\textbf{HSI-Drive}} & 1x1Conv & 81 & $\mathbf{1.03 \pm 0.04}$ & $\mathbf{0.12 \pm 0.01}$ \\
& AE & 1.9K & $2.76 \pm 0.19$ & $0.24 \pm 0.01$ \\
& ECA & 82 & $1.37 \pm 0.09$ & $0.24 \pm 0.01$ \\
& CBAM & 185 & $3.42 \pm 0.21$ & $3.99 \pm 0.59$ \\
& seAttn & 103 & $1.54 \pm 0.12$ & $0.22 \pm 0.01$ \\
& DSC & 334 & $2.33 \pm 0.11$ & $0.16 \pm 0.01$ \\
& ConvNeXt & 1.7K & $4.33 \pm 0.19$ & $0.27 \pm 0.01$ \\
\cmidrule{2-5}
& LQE$_1$ & \textbf{12} & $2.16 \pm 0.10$ & $1.00 \pm 0.03$ \\
& LQE$_3$ & \textbf{36} & $2.59 \pm 0.13$ & $0.96 \pm 0.77$ \\
\midrule
\multirow{9}{*}{\textbf{H-City}} & 1x1Conv & 390 & $12.86 \pm 0.82$ & $\mathbf{0.28 \pm 0.04}$ \\
& AE & 22K & $16.78 \pm 0.92$ & $0.74 \pm 0.02$ \\
& ECA & 391 & $\mathbf{11.45 \pm 0.84}$ & $0.31 \pm 0.01$ \\
& CBAM & 494 & $15.41 \pm 0.47$ & $9.50 \pm 0.45$ \\
& seAttn & 1.4K & $12.82 \pm 0.57$ & $0.34 \pm 0.05$ \\
& DSC & 1.7K & $24.51 \pm 0.30$ & $0.66 \pm 0.01$ \\
& ConvNeXt & 8.2K & $48.70 \pm 0.67$ & $2.02 \pm 0.24$ \\
\cmidrule{2-5}
& LQE$_1$ & \textbf{12} & $19.83 \pm 0.57$ & $1.25 \pm 0.15$ \\
& LQE$_3$ & \textbf{36} & $20.55 \pm 0.71$ & $1.81 \pm 0.09$ \\
\bottomrule
\multicolumn{5}{l}{\textsuperscript{*}~Inference time ($B=1$), averaged over 1K input hypercubes, with 10 warmup runs.}\\
\end{tabular}
}
\label{tab:computationPerformance}
\end{table}

\subsection{Computational Efficiency Analysis}
Table~\ref{tab:computationPerformance} presents computational metrics for all DR methods with spectral reduction to $F=3$:

\textbf{Parameters}. The proposed LQE maintains $4 \times P \times F$ trainable parameters, which remains constant regardless of input spectral dimensionality or spatial resolution, resulting in only 12 and 36 parameters for $P=1$ and $P=3$, respectively, as compared to 51--22K parameters for other learnable methods.

\textbf{Inference Speed}. LQE achieves GPU latency of 0.96--1.81~ms, which is slower than the fastest learnable baseline~(0.16--0.28 ms for 1$\times$1Conv) but remains competitive with relatively heavier architectures such as CBAM, DSC, and ConvNeXt. This latency overhead, however, is offset by LQE's parameter efficiency and consistent segmentation gains.


\subsection{Limitations and Future Directions}
\label{Suppl_Section_3:Limitation_FutureWork}

\subsubsection{Current Limitations}
\begin{itemize}
    \item \textit{Inference Latency}: LQE's inference time is challenging for ultra-low-latency applications requiring real-time video processing. Knowledge distillation to lightweight approximators and model pruning could address this limitation.
    \item \textit{Dataset-Specific Learned Filters}: Current LQE filters are optimized for individual datasets. While we observe consistent spectral patterns across SSM within each dataset (Fig.~\ref{fig:lqe_adaptability_SSMS}), cross-dataset transfer learning remains unexplored. This is due to: (1) largelly non-overlapping spectral ranges (HyKo-VIS:~470–630~nm vs. HSI-Drive:~600–975~nm), and (2) different scene statistics and annotations. Future work should investigate transfer learning within the overlapping spectral region between HSI-Drive and H-City~(600–950~nm).
\end{itemize}

\subsubsection{Future Research Directions}
\begin{itemize}
    
    \item \textit{Adaptive Peak Allocation}: Develop meta-learning strategies to automatically determine the optimal $P$ based on dataset spectral complexity and task requirements.

    \item \textit{Multi-Task Learning (Long-term)}: Extend LQE to jointly optimize for multiple perception tasks (segmentation, detection, depth estimation) within a unified spectral DR framework.
    
    \item \textit{Validation and Material-Aware Regularization}: Validation of LQE filters with optically realizable CFA prototypes to evaluate real-world performance. Additionally, we incorporate material-specific QE bounds (e.g., the response characteristics of silicon photodiodes) directly into the loss function to enforce manufacturability constraints during model optimization.
\end{itemize}

\section{Conclusion}
\label{sec:Conclusion}
This paper introduced LQE, a physics-inspired DR approach that parameterizes spectral response filter multi-peak asymmetric Gaussian basis functions to emulate realistic sensor QE curves. Extensive experiments across three urban driving HSI datasets and six SSMs demonstrate that LQE provides consistent performance over six conventional and seven learnable baseline DR methods while maintaining parametric efficiency. Across all configurations and datasets, LQE achieves average improvements of 1.25\%, 1.26\%, and 0.76\% in mIoU, mF1, and Kappa, respectively, while requiring only 12--36 trainable parameters compared to 51--22K for competing learnable approaches.
Ablation studies further show that low-order peak configurations provide an effective balance between model expressiveness and performance, delivering improved segmentation accuracy with minimal computational overhead (0.96--1.81~ms inference latency). The learned spectral response filter also exhibit interpretable, dataset-specific characteristics in urban driving scenes, highlighting the value of incorporating physics-informed inductive biases into DL architectures for hyperspectral perception.

This work provides a principled foundation for data-driven exploration of optimized multispectral sensor designs for real-world perception systems. Future work will focus on two directions: (1) incorporating manufacturability constraints to ensure practical sensor realizability, and (2) validating learned spectral responses through physical sensor prototypes to assess real-world feasibility. Such efforts will help bridge the gap between computational learning and physical sensor design, enabling next-generation hyperspectral sensing tailored for autonomous driving applications.


\bibliographystyle{IEEEtran}
\bibliography{ref}
\vspace{-1.7em}
\end{document}